%% file: 0_main.tex
\documentclass[11pt]{article}

\usepackage[preprint]{sty_acl/acl}

\usepackage{times}
\usepackage{latexsym}

\usepackage[T1]{fontenc}

\usepackage[utf8]{inputenc}

\usepackage{microtype}

\usepackage{inconsolata}

\usepackage{graphicx}

\usepackage{amsmath,amssymb,amsfonts}
\usepackage{algorithmic}
\definecolor{purple}{rgb}{0.51, 0.255, 0.51}
\definecolor{red}{rgb}{100, 0.0, 0}
\definecolor{green}{RGB}{0, 153, 76}
\definecolor{mag}{RGB}{204,0, 204}
\definecolor{darkblue}{RGB}{0, 0, 200}

\usepackage{textcomp}
\usepackage{url}
\usepackage{cleveref}
\usepackage{subcaption}
\usepackage{caption}
\usepackage{enumitem}
\input{math_commands}

\usepackage{booktabs}
\usepackage{multirow}
\usepackage[table]{xcolor}
\usepackage{graphicx}

\title{Neural Weight Compression for Language Models}

\author{
 \textbf{Jegwang Ryu\textsuperscript{1}},
 \textbf{Minkyu Kim\textsuperscript{1}},
 \textbf{Seungjun Shin\textsuperscript{2}},
 \\
 \textbf{Hee Min Choi\textsuperscript{2}},
 \textbf{Dokwan Oh\textsuperscript{2}},
 \textbf{Jaeho Lee\textsuperscript{1}}
\\
 \textsuperscript{1}POSTECH,
 \textsuperscript{2}Samsung Electronics Co., Ltd
\\
 \small{
   \textbf{Correspondence:} \href{mailto:jaeho.lee@postech.ac.kr}{jaeho.lee@postech.ac.kr}
 }
}

\begin{document}
\maketitle

\begin{abstract}
    \input{sections/0.abstract}
\end{abstract}

\input{sections/1.introduction}
\input{sections/2.framework}   
\input{sections/3.method}      
\input{sections/4.experiment}  
\input{sections/5.analyses}

\input{sections/6.related}
\input{sections/7.conclusion}  

\bibliography{ref}
\input{sections/99.appendix}

\end{document}

%% file: math_commands.tex
\usepackage{amsmath,amsfonts,bm}

\def\eqref#1{equation~\ref{#1}}

\def\1{\bm{1}}

\DeclareMathAlphabet{\mathsfit}{\encodingdefault}{\sfdefault}{m}{sl}
\SetMathAlphabet{\mathsfit}{bold}{\encodingdefault}{\sfdefault}{bx}{n}



%% file: sections/0.abstract.tex
Efficient compression of language model weights is increasingly critical as model scale and deployment grow. 
Yet, most existing methods rely on handcrafted transforms and heuristics, reflecting the limited understanding of \textit{weights} as a data modality. 
To move beyond this paradigm, we formulate weight compression as neural codec learning and propose Neural Weight Compression (NWC), a framework for training neural codecs on pretrained weight datasets. 
NWC addresses challenges intrinsic to weight compression, including tensor heterogeneity and the mismatch between reconstruction losses and downstream performance.
Experiments show that NWC achieves highly competitive accuracy--compression tradeoffs, with particularly strong results in the 4--6 bit regime, without relying on rigid handcrafted components such as the Hadamard transform. These gains extend to across diverse architectures, \textit{e.g.,} vision encoders. 
Our analysis highlights the roles of entropy-constrained quantization and learned transforms in adapting compression to weight data and downstream tasks.


%% file: sections/1.introduction.tex
\section{Introduction}\label{sec:intro}
The ``weights'' of neural nets constitute a new form of data, and the demand for efficient storage and transmission of this modality is rapidly increasing. This issue is particularly pressing for large language models (LLMs), whose parameter counts now reach hundreds of billions to the trillion scale \citep{gemini}. 
Beyond inference-time communication across intra- and inter-chip interconnects \citep{pope2023efficiently}, LLM weights must also sit in cold storage on public registries, exchanged during distributed or federated training \citep{mcmahan2017communication}, maintained as task- or user-specific updates for personalization \citep{hu2022lora}, and pile up as training-checkpoint archives whose footprint now rivals that of the training data itself. Therefore, effective weight compression is essential for reducing storage and transmission costs and making LLMs easier to deploy and distribute.

\input{figures/flow}

Currently, the dominant paradigm in compressing language model weights is post-training quantization with \textbf{\textit{lightweight handcrafted transforms}}, such as channel-wise scaling \citep{smoothquant, awq} and orthogonal rotation \citep{quarot, quip, qtip}. This design philosophy is driven by practical considerations: complex transformations can increase decoding overhead and reduce inference efficiency. However, restricting compression to manually designed transforms limits its ability to adapt to the highly heterogeneous statistics of model weights. Indeed, recent studies continue to propose alternative transforms \citep{van2025fptquant,liang2025paroquant}, suggesting that the community has yet to reach a consensus on what constitutes an optimal representation for compression.

This raises a fundamental question:
\begin{center}
\textit{``Can language model weight compression\\itself be \textbf{learned from data?}''}
\end{center}

This question is motivated by the success of neural codecs in natural-signal compression. For images, videos, and audio, learned codecs consistently outperform handcrafted designs by jointly learning representations and coding strategies directly from data \citep{balle, elic, hific}. Moreover, modern systems are increasingly memory-bound \citep{gholami2024ai}, and many deployment scenarios such as storage and transmission are considerably less sensitive to decoding latency, making richer compression schemes increasingly practical.


\paragraph{Challenges.} 
However, directly applying the idea of neural codecs to language model weights is nontrivial. Unlike natural signals, LLM weights exhibit several unique properties:
\begin{itemize}[leftmargin=*,topsep=0pt,parsep=0pt,itemsep=0pt]
\item \textit{Heterogeneity of weight tensors}: Weight tensors vary substantially in shape and scale, exhibiting diverse dimensionalities and statistics.
\item \textit{Downstream performance beyond MSE}: The quality of reconstructed weights should be assessed by their impact on downstream model performance, rather than by element-wise reconstruction error alone.
\item \textit{Structured outliers in weights}: Unlike natural signals, LLM weight tensors often contain pronounced and structured outliers.
\end{itemize}

\paragraph{Contribution.} 
By addressing these challenges, we develop Neural Weight Compression (NWC), a neural codec for LLM weights based on the nonlinear transform-coding paradigm of \citet{balle}.
Specifically, our framework introduces three key components: (i) chunk-and-normalize preprocessing, (ii) an importance-aware training loss that prioritizes chunks more critical to model performance; and (iii) inference-time error compensation that propagates errors through each layer.

Without relying on any handcrafted transforms, NWC achieves competitive accuracy-compression tradeoffs.
This benefit is particularly strong in the 4--6 bit regime on language models and extends to vision encoders for LLMs.
Our analyses suggest that this success may be due to both the inherent limitations of the competing VQ-based baselines in handling high bitrates \citep{qtip}, and the ability of learned transform coding to capture features that are relevant to the downstream task.

To sum up, NWC presents a flexible yet effective approach for compressing the weights. Our work provides a solid starting point for a fully automated compression pipeline for model weights.

%% file: figures/flow.tex

\begin{figure*}[t!]
    \centering
    \begin{subfigure}[b]{1\textwidth}
        \includegraphics[width=\textwidth]{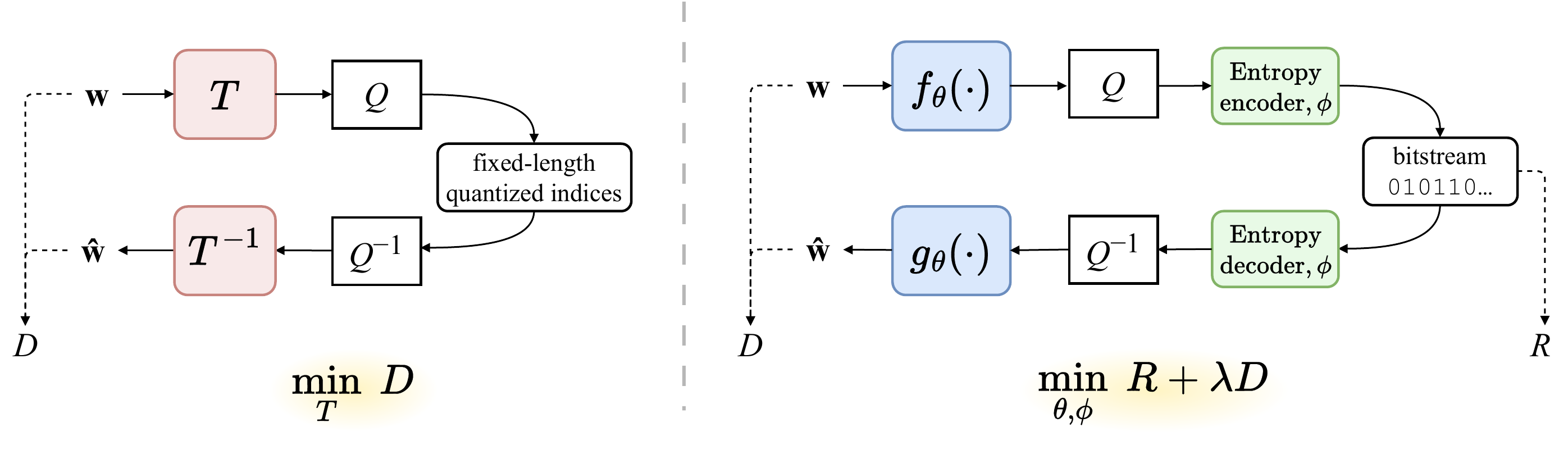}
        \label{fig:flow_quant}
    \end{subfigure}
    \vspace{-3.5em}
    \caption{
    Two weight compression paradigms and their optimization objectives.
    \textbf{(Left)} \textbf{Conventional weight compression} uses a linear transform, mapping $\mathbf{w} \mapsto T\mathbf{w}$ before quantization and inverse mapping $\mathbf{z} \mapsto T^{-1}\mathbf{z}$. With the predetermined rate, the objective reduces to minimizing the distortion $D$. 
    \textbf{(Right)} \textbf{The proposed neural weight compression} replaces the transform with learned nonlinear analysis and synthesis transforms, $f_{\theta}$ and $g_{\theta}$, and uses an entropy model $\phi$ for variable-length coding. The codec is optimized under the rate--distortion objective.
    }
    \label{fig:overview}
\end{figure*}

%% file: sections/2.framework.tex
\input{figure_tex/architecture}

\section{Problem formulation}

Consider a weight codec mapping the weight $\mathbf{w} \in \mathbb{R}^d$ to a reconstruction $\hat{\mathbf{w}} \in \mathbb{R}^d$ through a finite bitstream. 
Let $f: \mathbb{R}^d \rightarrow \mathbb{R}^k$ be an \textit{analysis transform} that maps $\mathbf{w}$ to a latent representation $\mathbf{z} \in \mathbb{R}^k$, and let $g: \mathbb{R}^k \rightarrow \mathbb{R}^d$ be a \textit{synthesis transform} that maps the latent representation back to the weight space. 
Then, the compression and decompression processes can be written as:
\begin{align}
\text{Compression:}& \quad 
\mathbf{b} 
= C(Q(f(\mathbf{w}))), 
\label{eq:compression} \\[0.4em]
\text{Decompression:}& \quad 
\hat{\mathbf{w}} 
= g(Q^{-1}(C^{-1}(\mathbf{b}))),
\label{eq:decompression}
\end{align}
where the quantizer $Q$ discretizes the latent representation, and the coder $C$ encodes the quantized values into a finite length bitstream $\mathbf{b} \in \{0, 1\}^\ast$.

Our goal is to minimize the distortion on $\mathbf{w}$ given the rate constraints. Formally, let 
$d: \mathbb{R}^d \times \mathbb{R}^d \rightarrow \mathbb{R}$ be a task-dependent distortion measure. We consider the constrained optimization:
\begin{gather}
    \min_{f, g, C} \quad 
    \mathbb{E}_{\mathbf{w} \sim \mathcal{D}_{\mathcal{W}}}
    [d(\mathbf{w}, \hat{\mathbf{w}})], \\
    \text{subject to} \quad
    \mathbb{E}_{\mathbf{w}}[\mathrm{len}(\mathbf{b})] \le R,
\end{gather}
where $\mathrm{len}(\cdot)$ denotes the bitstream length, and $R$ is the rate constraint imposed.

Unlike conventional codecs, where distortion is typically a reconstruction error (e.g., MSE), weight compression should account for downstream performance. 
Let $\mathcal{M}_{\mathbf{w}}$ and $\mathcal{M}_{\hat{\mathbf{w}}}$ denote the models parameterized by the original and reconstructed weights, respectively. 
The distortion is defined as
\begin{align}
d(\mathbf{w}, \hat{\mathbf{w}}) =  \mathrm{Err}_{\mathcal{T}}
    (\mathcal{M}_{\mathbf{w}}, \mathcal{M}_{\hat{\mathbf{w}}}),\label{eq:task_error}
\end{align}
where $\mathrm{Err}_{\mathcal{T}}$ is a task-dependent error functional that measures the performance degradation caused by replacing $\mathbf{w}$ with $\hat{\mathbf{w}}$ on a downstream task $\mathcal{T}$.

\subsection{Transforms for weight compression}

The presence of outliers within LLM weights has motivated the use of transforms tailored to weight compression. 
Prior arts commonly apply invertible linear transforms before quantization, including \textit{channel scaling} \citep{awq, smoothquant} and \textit{rotation} \citep{quarot, quip, liang2025paroquant}. Specifically, given the weight $\mathbf{w}$, the transforms are formulated as:
\begin{align}
    f: \mathbf{w} \mapsto T\mathbf{w},
    \qquad 
    g: \mathbf{z} \mapsto T^{-1}\mathbf{z}.
\end{align}
In \textit{channel scaling}, $T$ is a diagonal matrix. 
In \textit{rotation}, $T$ is an orthogonal matrix, so that $T^{-1} = T^\top$.

\subsection{Learned weight compression}

We consider learned compression with nonlinear transforms and entropy coding. We parameterize the \textit{analysis} and \textit{synthesis} transforms as neural nets:
\begin{align}
    f = f_{\theta_a}(\cdot),
    \quad
    g = g_{\theta_s}(\cdot),
\end{align}
and optimize them jointly with a learnable entropy model, $p_{\phi}$. 
The learning objective is the Lagrangian relaxation of the rate--distortion problem:
\begin{align}
&\min_{\theta_a,\theta_s,\phi} \quad
\mathbb{E}_{\mathbf{w}}
\left[
-\log p_{\phi}(\hat{\mathbf{z}})
+
\lambda d(\mathbf{w}, \hat{\mathbf{w}})
\right],
\label{eq:learned_rd_objective}\\
&\mathrm{where}\quad\hat{\mathbf{z}} = Q(f_{\theta_a}(\mathbf{w})),\quad \hat{\mathbf{w}} = g_{\theta_s}(\hat{\mathbf{z}}) \nonumber
\end{align}
Here, $\mathbb{E}[-\log p_{\phi}(\hat{\mathbf{z}})]$ denotes the rate loss with respect to the learned entropy, which penalizes the expected code length and the parameter $\lambda$ controls the trade-off between this rate and the distortion. To circumvent the non-differentiability of quantization during training, additive uniform noise is employed as a relaxation \citep{balle}.

This formulation allows the codec to learn both the representation and the code length directly from weight data, rather than relying on a manually designed transform and a fixed bitwidth.

%% file: figure_tex/architecture.tex
\begin{figure*}[t!]
    \centering
    \includegraphics[width=0.9\textwidth]{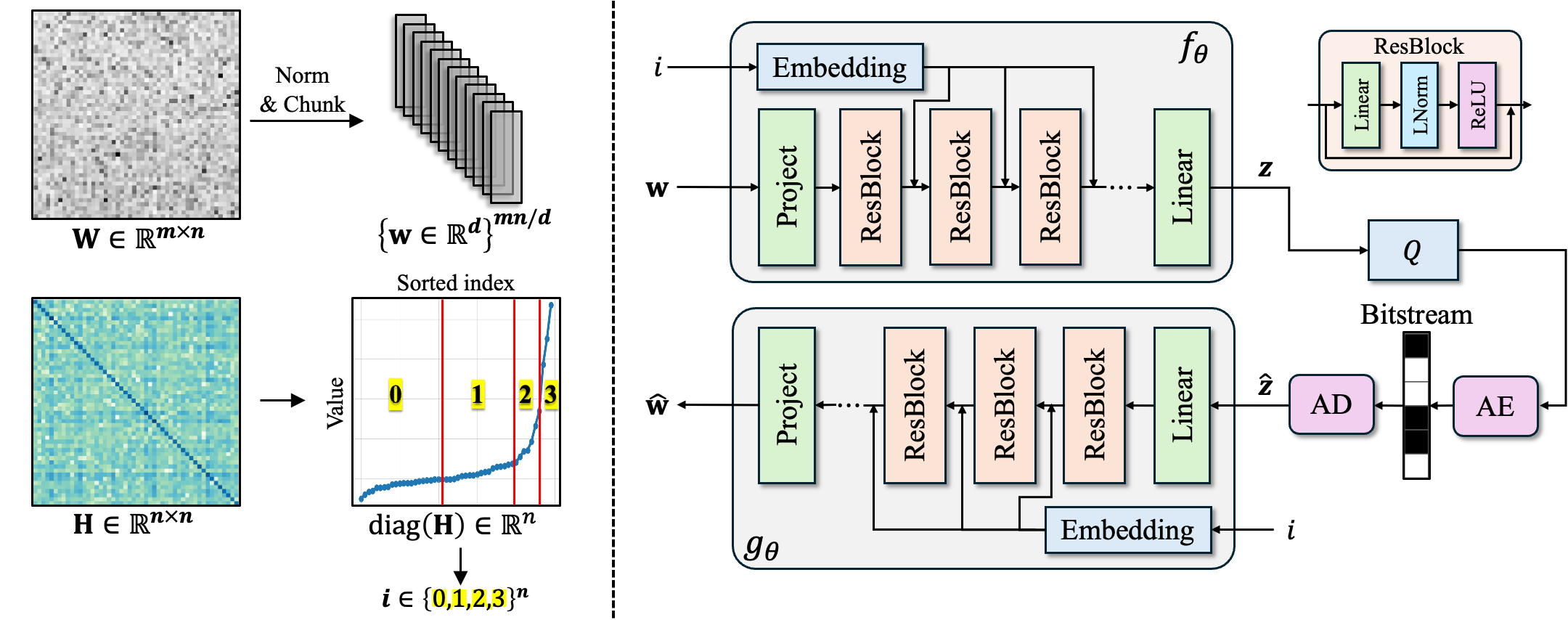}
    \vspace{-1em}
    \caption{The proposed neural weight compression (NWC) framework. (Left) Preprocessing steps for weight tensors, including column-wise chunk-and-normalization and importance level assignment. (Right) Model architectures of the analysis and synthesis. AE/AD refer to Arithmetic Encoding/Decoding. $Q$ denotes the rounding operator.}
    \label{fig:archi}
\end{figure*}

%% file: sections/3.method.tex
\section{Neural weight compression}\label{sec:method}

Now we describe the proposed neural weight compression (NWC) framework for LLM weights.

\subsection{Preprocessing: Chunk and normalize}\label{ssec:preprocessing}

Unlike images or videos, LLM weights vary widely in shape and scale. For instance, in Llama 3-8B, the key-projection matrix has size $\mathbb{R}^{1024 \times 4096}$, whereas the up-projection matrix has size $\mathbb{R}^{14336 \times 4096}$. Their statistics also differ substantially across layers and channels; see \Cref{ssec:per_layer_scale,ssec:channel_scale}.

To handle this heterogeneity, we preprocess each weight matrix $\mathbf{W}\in\mathbb{R}^{m\times n}$ with a simple column-wise pipeline. We first split $\mathbf{W}$ into column vectors 
$\mathbf{w}_{\mathrm{col}}\in\mathbb{R}^{m}$, normalize each column to unit standard deviation, and then divide them into fixed-length chunks 
$\mathbf{w}\in\mathbb{R}^{16}$, which serve as inputs to the neural codec; see \Cref{fig:archi}, left.

Column-wise normalization factors are stored in FP16 and used to restore the original scale after reconstruction. This incurs only about 0.004 bits per parameter. We use column-wise chunking because it aligns naturally with our inference-time error compensation procedure, described in \Cref{ssec:inference}.

\subsection{Training}\label{ssec:training}

Ideally, the codec should be trained to directly minimize $\mathrm{Err}_{\mathcal{T}}(\mathcal{M}_{\mathbf{w}}, \mathcal{M}_{\hat{\mathbf{w}}})$. For LLMs, however, this is impractical: Each update would require decoding the weights, running the full model on the task, and attributing the resulting error to individual compressed chunks. We therefore adopt a local proxy based on the output distortion of each linear layer.

For a linear layer $\mathbf{x} \mapsto \mathbf{Wx}$, the effect of replacing $\mathbf{W}$ with $\hat{\mathbf{W}}$ can be measured as
\begin{align}
&\mathbb{E}[\|\mathbf{Wx}-\hat{\mathbf{W}}\mathbf{x}\|^2_2] \nonumber\\
&= \mathrm{tr}\left((\mathbf{W}-\hat{\mathbf{W}})\mathbf{H}(\mathbf{W}-\hat{\mathbf{W}})^\top\right)
\end{align}
where the Hessian $\mathbf{H} = \mathbb{E}[\mathbf{xx}^\top]$ is estimated from calibration activations. 
Motivated by activation outliers in LLMs \citep{an2025systematic,sun2024massive}, we use the diagonal approximation
\begin{align}
\approx \mathrm{tr}\left((\mathbf{W}-\hat{\mathbf{W}})\mathrm{diag}(\mathbf{H})(\mathbf{W}-\hat{\mathbf{W}})^\top\right),
\end{align}
which yields a Hessian-weighted MSE: errors in columns with larger activation (or Hessian diagonals) are penalized more heavily.

\paragraph{Chunk-wise importance weights.} Since NWC compresses fixed-length chunks for each column, all chunks from the same column inherit the corresponding Hessian diagonal. We discretize these values into $K$ importance levels $i \in \{0,\ldots,K-1\}$ and assign each level a weight $\lambda_I^{(i)}$. The training objective for a chunk $\mathbf{w}$ is:
\begin{align}
\mathcal{L}_{\mathrm{Imp}} = \mathbb{E}[-\log p_{\phi}(\hat{\mathbf{z}}) +\lambda \lambda_I^{(i)} \|\mathbf{w}-\hat{\mathbf{w}}\|^2_2].
\end{align}
Here, $\lambda_I^{(i)}$ controls the reconstruction fidelity allocated to chunks of importance level $i$. In practice, a small number of levels, e.g., $K=4$, is sufficient.

\paragraph{Randomized importance conditioning.} During training, we sample the importance level $i$ uniformly at random and condition the codec on it. At inference, we instead use the Hessian-derived level of each column; see \Cref{fig:archi} (left) and  \Cref{sec:importance_detail}. This allows a single codec to support multiple rate-distortion tradeoffs, reconstructing sensitive chunks more accurately while compressing less sensitive chunks more aggressively.

\paragraph{Network architectures.} As shown in the right panel of \Cref{fig:archi}, both the analysis transform $f$ and synthesis transform $g$ are residual MLPs. The importance index $i$ is embedded and injected into each residual block by element-wise multiplication with the hidden states. Storing this index requires only $\lceil \log_2(K) \rceil$ bits per column, adding $<0.001$ bits per parameter. We use a fully factorized entropy model with arithmetic coding \citep{balle}.

\input{figure_tex/8b_wiki_mmlu_common}
\input{figure_tex/result_other_models_LLM}

\subsection{Inference}\label{ssec:inference}
At inference time, we compress model weights sequentially. To reduce error accumulation, we compensate for residuals from previously compressed weights by absorbing them into the remaining uncompressed weights.

\noindent\textbf{Intra-layer error compensation.}
When compressing a weight matrix column by column, we adjust each uncompressed column using the residuals from earlier compressed columns. Let $\mathbf{r}_i=\mathbf{w}_i-\hat{\mathbf{w}}_i$ denote the reconstruction residual of the $i$-th column. Before compressing the $k$-th column, we construct an error-compensated column as
\begin{align}
\tilde{\mathbf{w}}_k
&= \mathbf{w}_k+\sum_{i<k} \mathbf{r}_i c_{i,k} \\
&= \mathbf{w}_k + (\mathbf{W}_{1:k-1}-\hat{\mathbf{W}}_{1:k-1})\mathbf{c}_k ,
\end{align}
where $\mathbf{c}_k$ contains the feedback coefficients from previous columns to the current one. Following prior work \citep{quip}, we set $\mathbf{c}_k$ to the corresponding off-diagonal entries of the $k$-th column of $\mathbf{L}$ from the LDL decomposition of the layer Hessian, $\mathbf{H} = \mathbf{L}^\top \mathbf{D}\mathbf{L}$. Our column-wise chunking makes the codec compatible with this procedure.

\paragraph{Inter-layer recovery fine-tuning.} We also compensate for errors across layers within each transformer block. Before compressing a layer, we fine-tune the remaining uncompressed layers in the same block to account for already compressed layers. After each layer in the $k$-th block is compressed, we optimize the remaining layers to minimize the MSE between the current block output and the original uncompressed block output.

As block inputs, we use the calibration features computed by the original model. This enables different transformer blocks to be compressed in parallel, as in \citet{quip,aqlm}

%% file: figure_tex/8b_wiki_mmlu_common.tex
\begin{figure*}[t!]
    \centering
    \begin{subfigure}[b]{0.32\textwidth}
        \includegraphics[width=\textwidth]{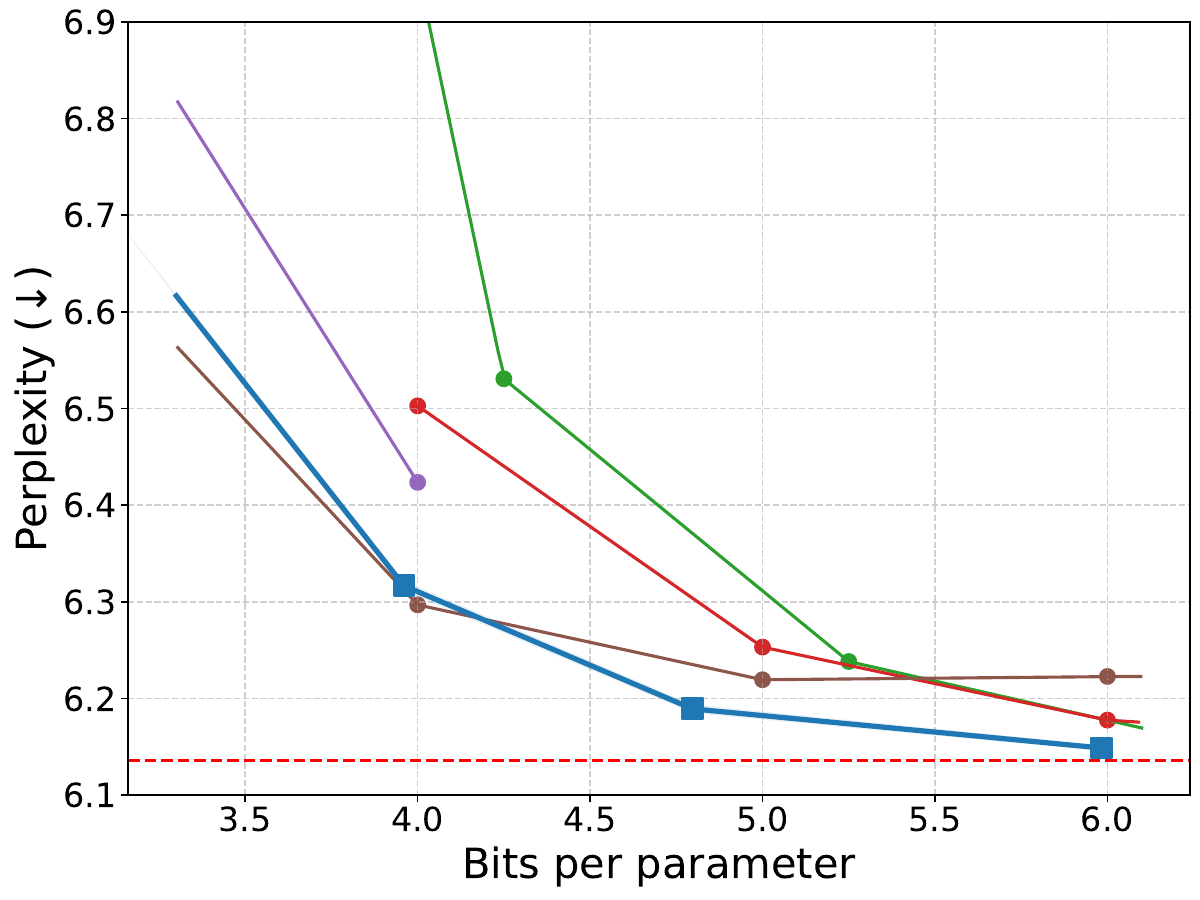}
        \caption{WikiText-2}
        \label{fig:8b_c4}
    \end{subfigure}
    \begin{subfigure}[b]{0.32\textwidth}
        \includegraphics[width=\textwidth]{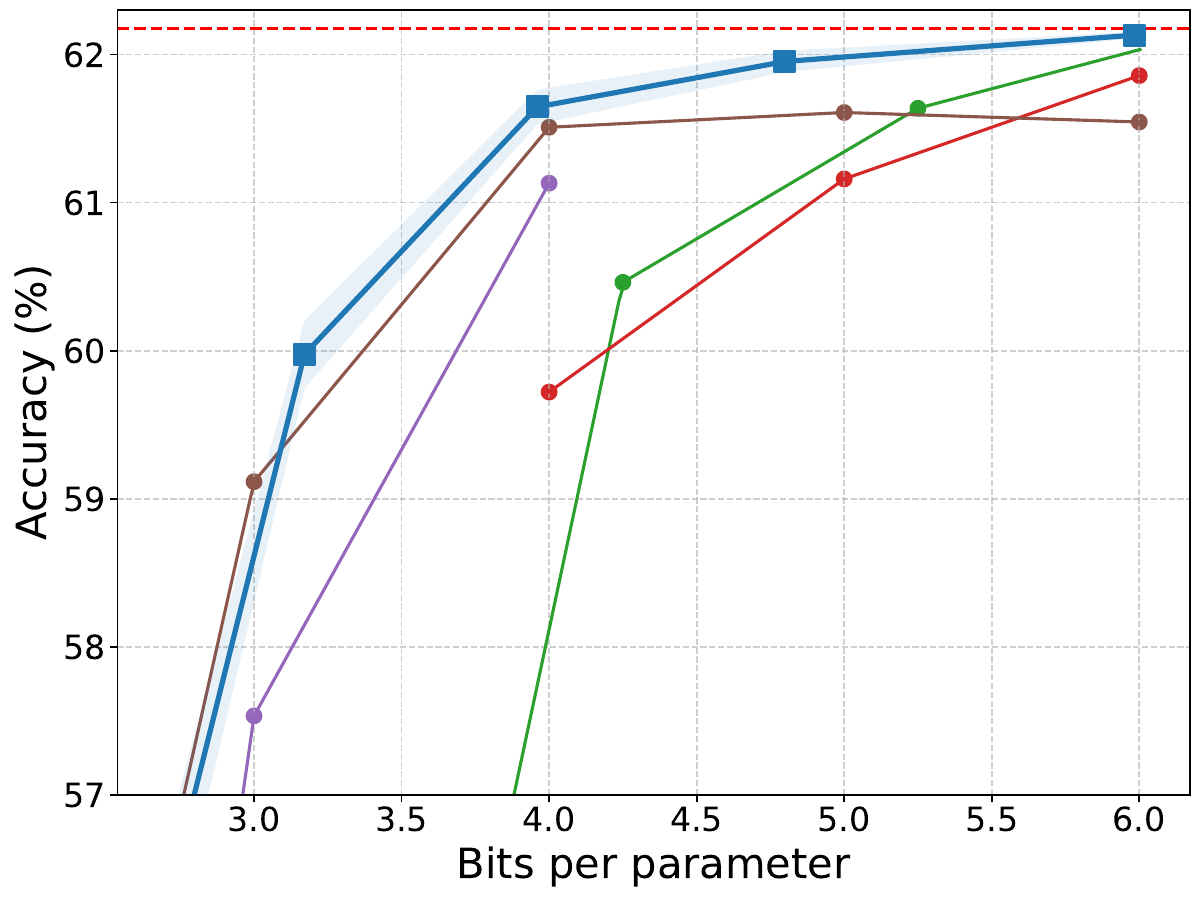}
        \caption{MMLU}
        \label{fig:7b_c4}
    \end{subfigure}
    \begin{subfigure}[b]{0.32\textwidth}
        \includegraphics[width=\textwidth]{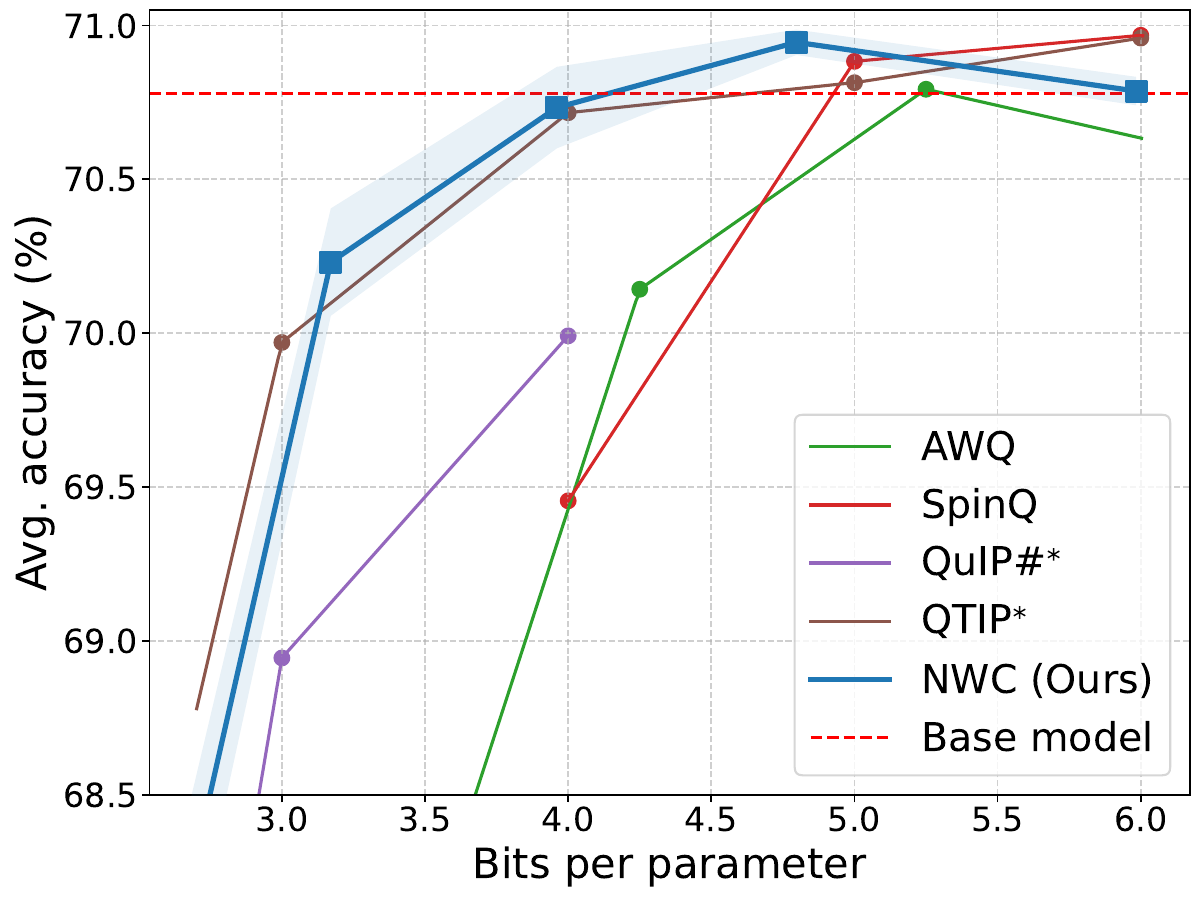}
        \caption{Commonsense tasks}
        \label{fig:13b_c4}
    \end{subfigure}
    \caption{Rate-accuracy tradeoffs on Llama 3-8B. We evaluate WikiText-2 perplexity with a context length of 2048, as well as zero-shot accuracies on MMLU and six common-sense tasks---ARC-Easy, ARC-Challenge, WinoGrande, PiQA, HellaSwag, BoolQ---across varying average bit-widths. NWC results are reported as an average over three random seeds, and the standard error is marked in shade.}
    \label{fig:8b_wiki_mmlu_common}
\end{figure*}

%% file: figure_tex/result_other_models_LLM.tex
\begin{figure*}[ht!]
    \centering
    \hfill
    \begin{subfigure}[b]{0.32\textwidth}
        \includegraphics[width=\textwidth]{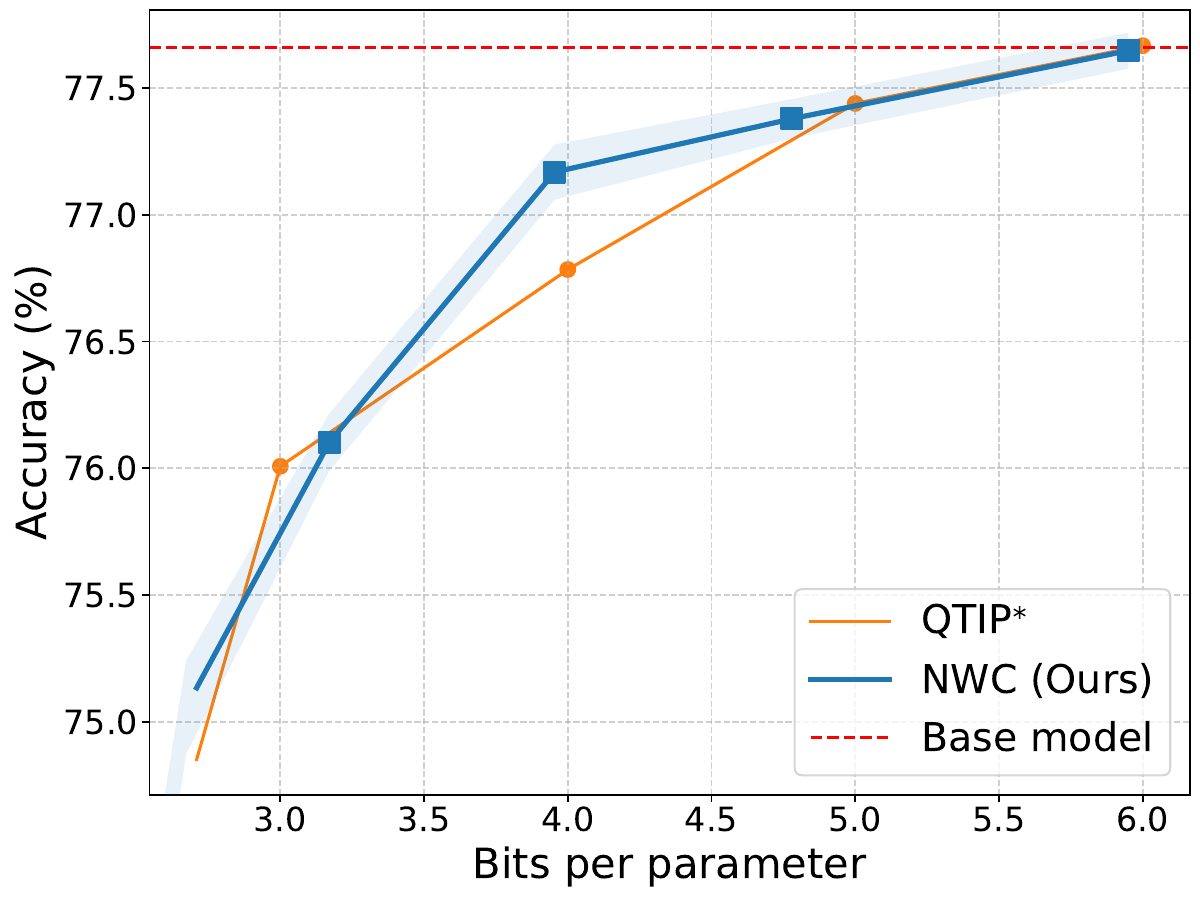}
        \caption{Qwen 3-30B-A3B}
    \end{subfigure}
    \hfill
    \begin{subfigure}[b]{0.32\textwidth}
        \includegraphics[width=\textwidth]{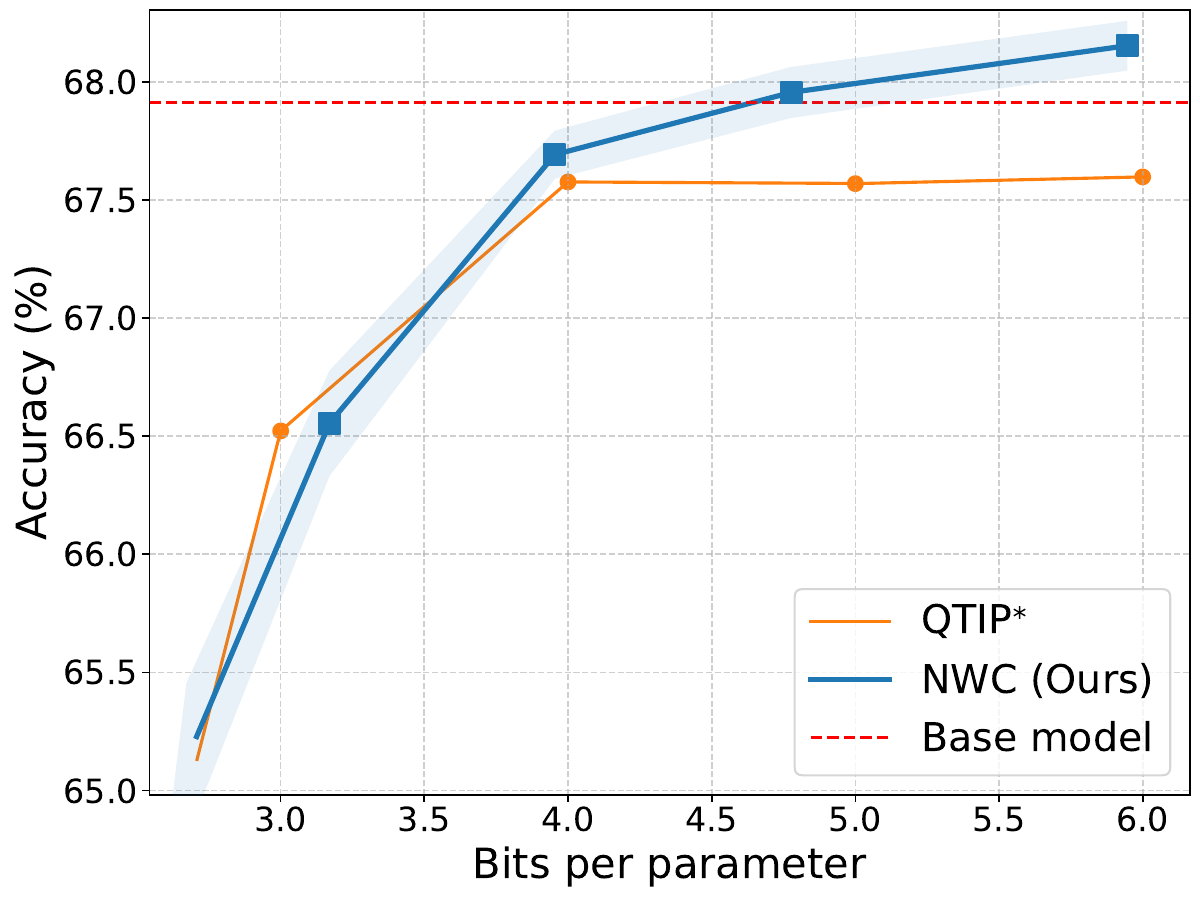}
        \caption{Mixtral-8x7B}
    \end{subfigure}
    \hfill
    \begin{subfigure}[b]{0.32\textwidth}
        \includegraphics[width=\textwidth]{"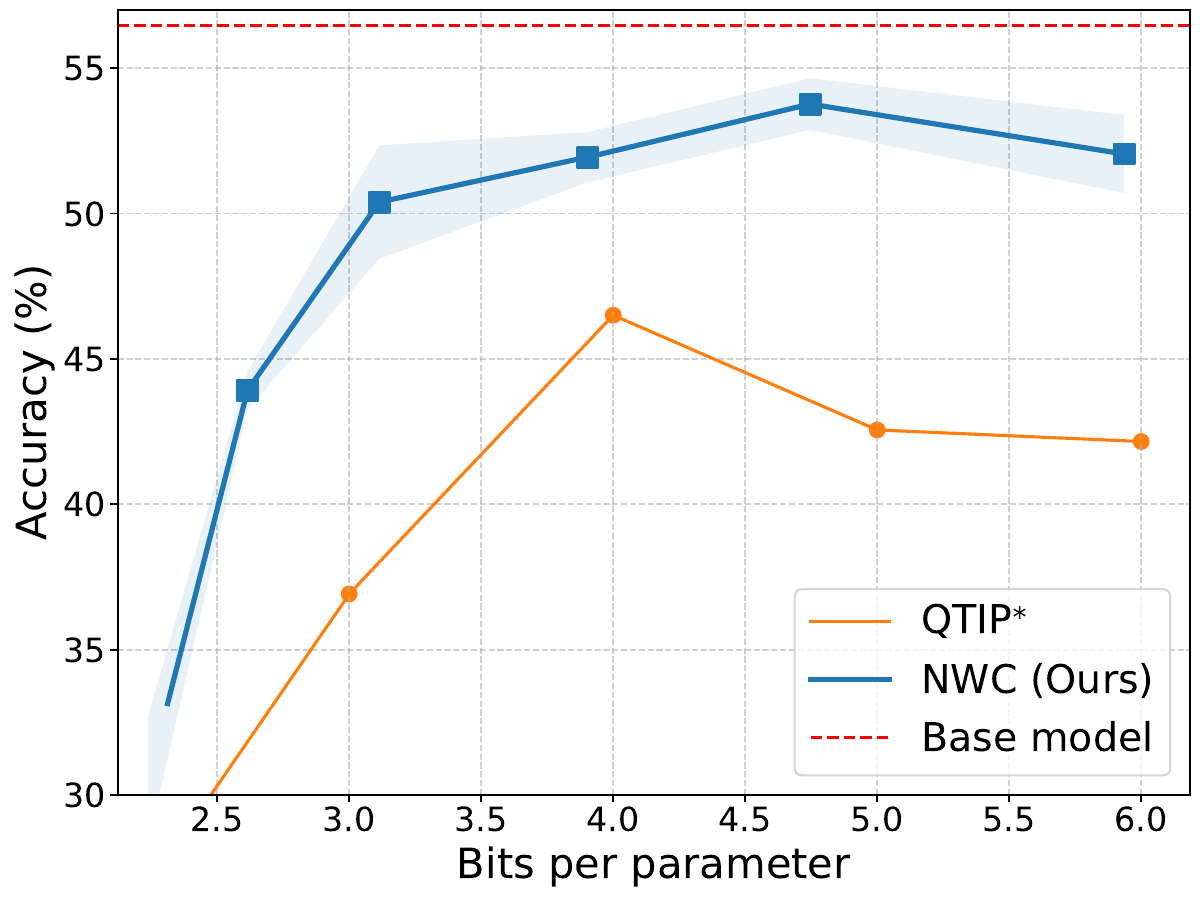"}
        \caption{GPT-OSS-20B}
    \end{subfigure}
    \hfill
    \caption{Rate-accuracy tradeoff across diverse LLM architectures---Qwen, Mixtral, and GPT-OSS---on MMLU. NWC results are reported as an average over three random seeds, and the standard error is marked in shade.}
    \label{fig:other_models}
    \vspace{-1em}
\end{figure*}

%% file: sections/4.experiment.tex
\section{Experiments}\label{sec:experiments}

\subsection{Experimental setup}
\paragraph{Baselines.} We compare NWC against post-training model compression methods, including (1) Scalar PTQ: AWQ \citep{awq}, GPTQ \citep{gptq}, and SpinQuant \citep{spinquant}; (2) Vector PTQ: QuIP\# \citep{quip-sharp} and QTIP \citep{qtip}; (3) Pseudo-random generator: SeedLM \citep{seedlm}; {(4) Neural codec: ReALLM \citep{reallm}}. 

Comparisons with alternative baselines such as QAT, SVD-based methods, and PocketLLM \citep{tian2025pocketllm} are provided in \Cref{ssec:nonptq}.

To assess the efficacy of the compression scheme in isolation, we do not conduct end-to-end fine-tuning in any methods compared. To clarify this point, we mark the modified baselines---QuIP\# and QTIP---with $^*$. See \Cref{ssec:setup_details} for details. 


\paragraph{Evaluation.} We measure the compression quality with three types of metrics. 
(1) Perplexity: WikiText2~\citep{wikitext}, and C4~\citep{c4}; 
(2) Zero-shot accuracy: MMLU \citep{mmlu}, and 6 common-sense tasks (ARC-Easy, ARC-Challenge, WinoGrande, PiQA, HellaSwag, BoolQ); 
(3) Reasoning accuracy: MMLU-Pro~\citep{wang2024mmlu}, GPQA Diamond~\citep{rein2023gpqa}, and AIME~\citep{maa2025aime}.

\paragraph{Training.} We train the codec on a dataset consisting of all linear layer weight tensors from Llama 3-8B. 
Both encoder and decoder consist of 4-layer residual MLPs with a width of 512. 
See \Cref{sec:network_design} for more details.

\subsection{LLM weight compression}\label{sec:llm}

\Cref{fig:8b_wiki_mmlu_common} reports the quality metrics of compressed Llama 3 \citep{llama3} at various rates; see \Cref{fig:llama_c4} for C4 perplexity. 
NWC consistently outperforms most baselines, achieving better tradeoff. Notably, the advantage of the neural approach becomes more pronounced at rates over 4 bits.

\input{figure_tex/result_vision_models}
\input{tables/longbench}
\input{figure_tex/datafree}

\paragraph{Generalization to diverse architectures.}
Without retraining the codec, we apply the framework to the diverse LLM architectures, including Qwen 3 \citep{yang2025qwen3}, Mixtral \citep{jiang2024mixtral}, and GPT-OSS \citep{agarwal2025gpt}. In \Cref{fig:other_models}, we observe that the performance on these models is strong in the 4--6 bit regime, similar to what has been observed on the Llama models.


\paragraph{Reasoning tasks.} 
\Cref{tab:qwen3_4bit} compares performance on Qwen3-8B and Qwen3-4B across reasoning benchmarks at 4-bit. We observe that NWC consistently matches the accuracy of the prior art with handcrafted transforms, showing that it effectively preserves the long generation capacity of the full-precision base models.

\subsection{Data-free setting}
In \Cref{fig:seedlm}, we compare the performance of NWC with ReALLM, SeedLM, and QTIP in a setting without calibration data. In this scenario, NWC and QTIP do not use Hessian-based error compensation (LDLQ) and do not apply fine-tuning. NWC consistently achieves lower perplexity compared to these baselines across various bitrates.

\subsection{Vision encoders}
In \Cref{fig:vision_models}, we evaluate NWC on the prominent vision encoder, including CLIP-ViT-L/16~\citep{clip}, SigLIP-B/16~\citep{siglip} and DINOv2-L~\citep{dinov2} on ImageNet~\citep{imagenet}. The results show that NWC achieves superior performance at mid-to-high bitrates. This is consistent with the trend observed in the LLM experiments, suggesting that the benefits of our approach generalize across the neural networks trained on data from different modalities. See \Cref{sec:vision_details} for more setup details.

\subsection{Other experiments}
In \Cref{sec:additional,sec:ablation}, we provide additional experimental results on the following topics:
\begin{itemize}[leftmargin=*,topsep=0pt,parsep=0pt,itemsep=0pt]
    \item Additional benchmarks (\Cref{ssec:additional_llm})
    \item Additional baselines (\Cref{ssec:nonptq})
    \item Ablation studies (\Cref{sec:ablation})
\end{itemize}

%% file: figure_tex/result_vision_models.tex
\begin{figure*}[ht!]
    \centering
    \hfill
    \begin{subfigure}[b]{0.32\textwidth}
        \includegraphics[width=\textwidth]{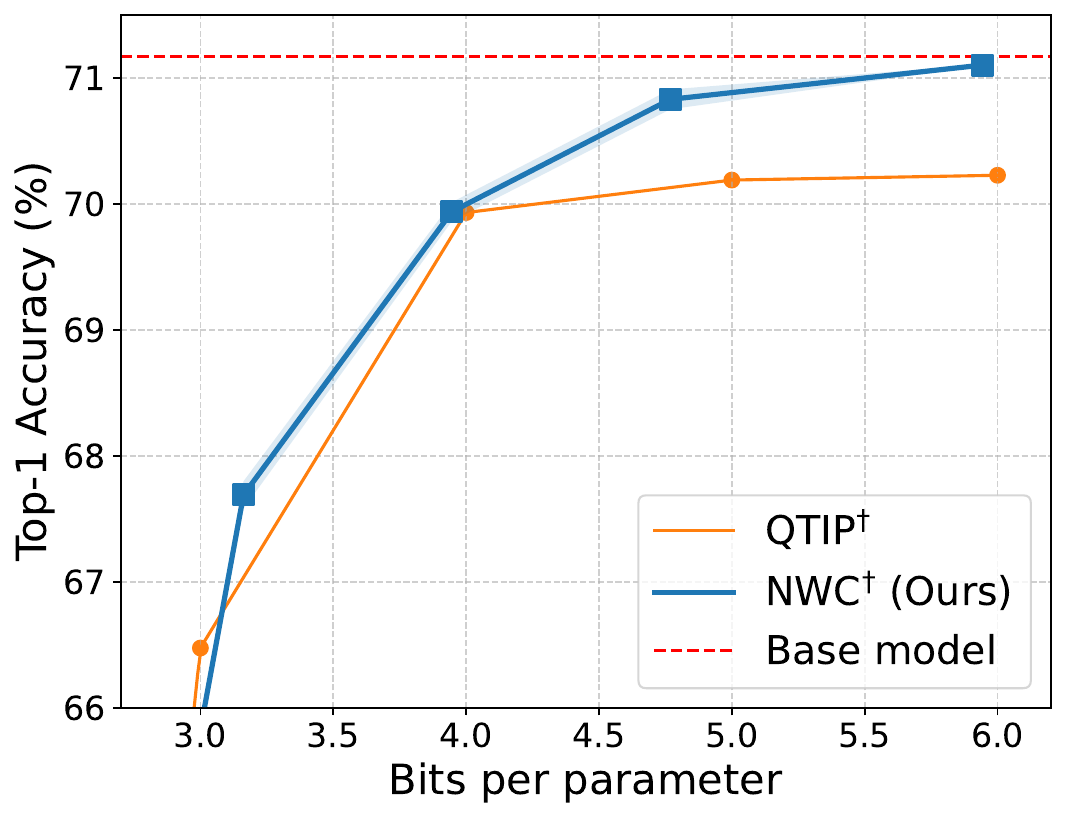}
        \caption{CLIP}
    \end{subfigure}
    \hfill
    \begin{subfigure}[b]{0.32\textwidth}
        \includegraphics[width=\textwidth]{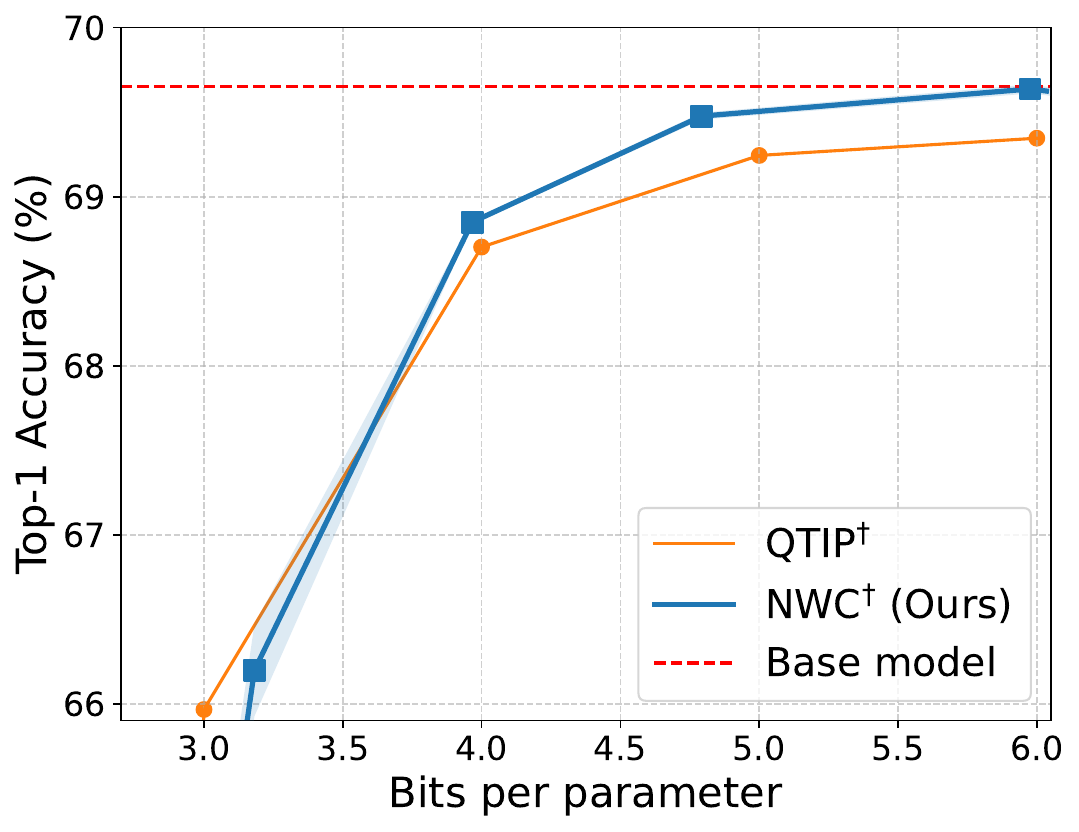}
        \caption{SigLIP}
    \end{subfigure}
    \hfill
    \begin{subfigure}[b]{0.32\textwidth}
        \includegraphics[width=\textwidth]{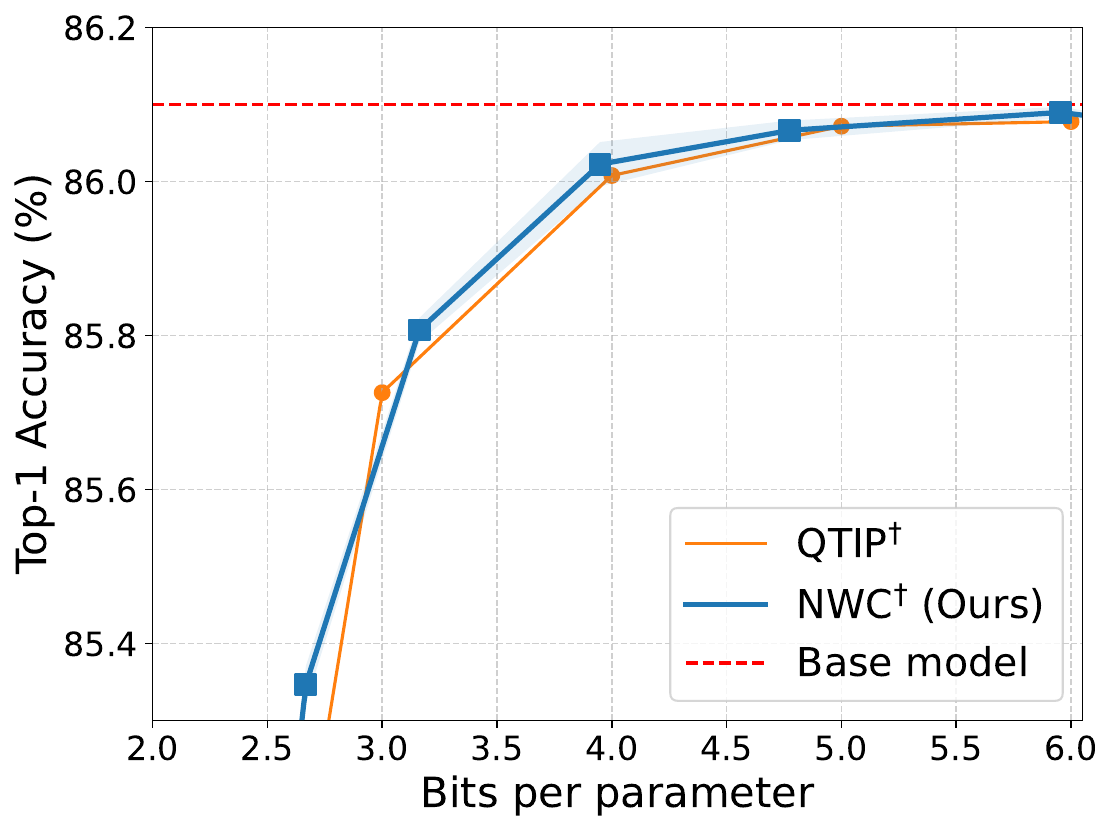}
        \caption{DINOv2}
    \end{subfigure}
    \hfill
    \caption{Compression of vision encoders. We report zero-shot classification accuracy for CLIP and SigLIP, and linear probing accuracy for DINOv2 on ImageNet-1k.}
    \label{fig:vision_models}
    \vspace{-1em}
\end{figure*}

%% file: tables/longbench.tex
\begin{table}[t]
\centering
\resizebox{\linewidth}{!}{%
\begin{tabular}{lcccccc}
\toprule
\textbf{Method} & \textbf{Bit} & \textbf{MMLU-Pro} & \textbf{GPQA} & \textbf{AIME-24} & \textbf{AIME-25} & \textbf{Avg.} \\
\midrule

\multicolumn{7}{l}{\textit{Qwen3-8B}} \\
Base  & 16   & 74.8          & 58.6          & 73.3          & 73.3          & 70.0 \\
QTIP  & 4    & \textbf{74.0} & 57.7          & 70.0          & 68.9          & 67.7 \\
NWC   & 3.94 & 73.8          & \textbf{58.8} & \textbf{71.1} & \textbf{71.2} & \textbf{69.0} \\
\midrule
\multicolumn{7}{l}{\textit{Qwen3-4B}} \\
Base  & 16   & 70.7          & 54.0          & 73.3          & 60.0          & 64.5          \\
QTIP  & 4    & \textbf{69.8} & \textbf{55.2} & 71.1          & 57.8          & 63.5          \\
NWC   & 3.94 & 69.4          & 53.2          & \textbf{73.3} & \textbf{61.1} & \textbf{63.7} \\
\bottomrule
\end{tabular}
}
\caption{Reasoning task accuracy comparison at 4-bit compression on Qwen3-8B and Qwen3-4B.}\label{tab:qwen3_4bit}
\end{table}



%% file: figure_tex/datafree.tex
\begin{figure}[t]
    \centering
    \begin{subfigure}[b]{0.49\columnwidth}
        \includegraphics[width=\textwidth]{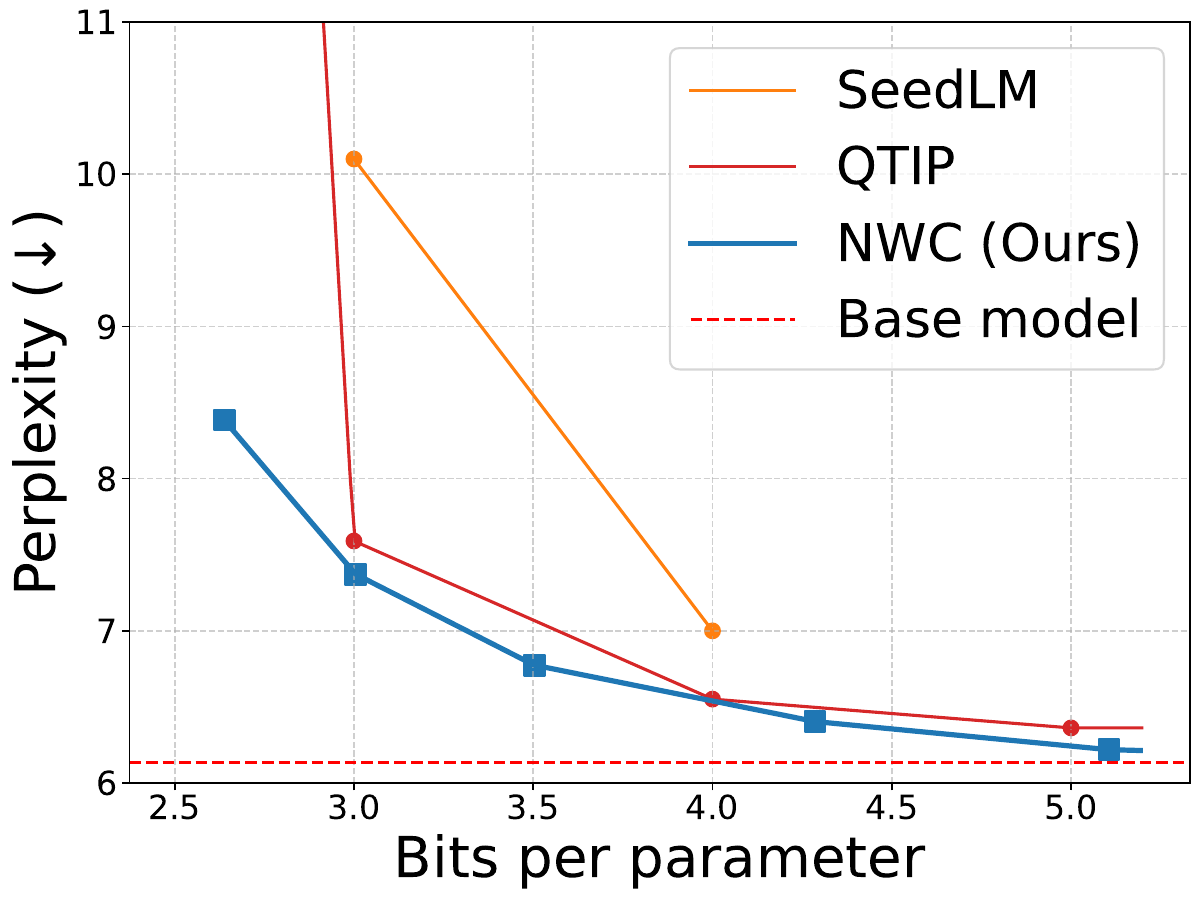}
        \caption{Llama 3-8B}
        \label{fig:8b_seedlm}
    \end{subfigure}
    \begin{subfigure}[b]{0.49\columnwidth}
        \includegraphics[width=\textwidth]{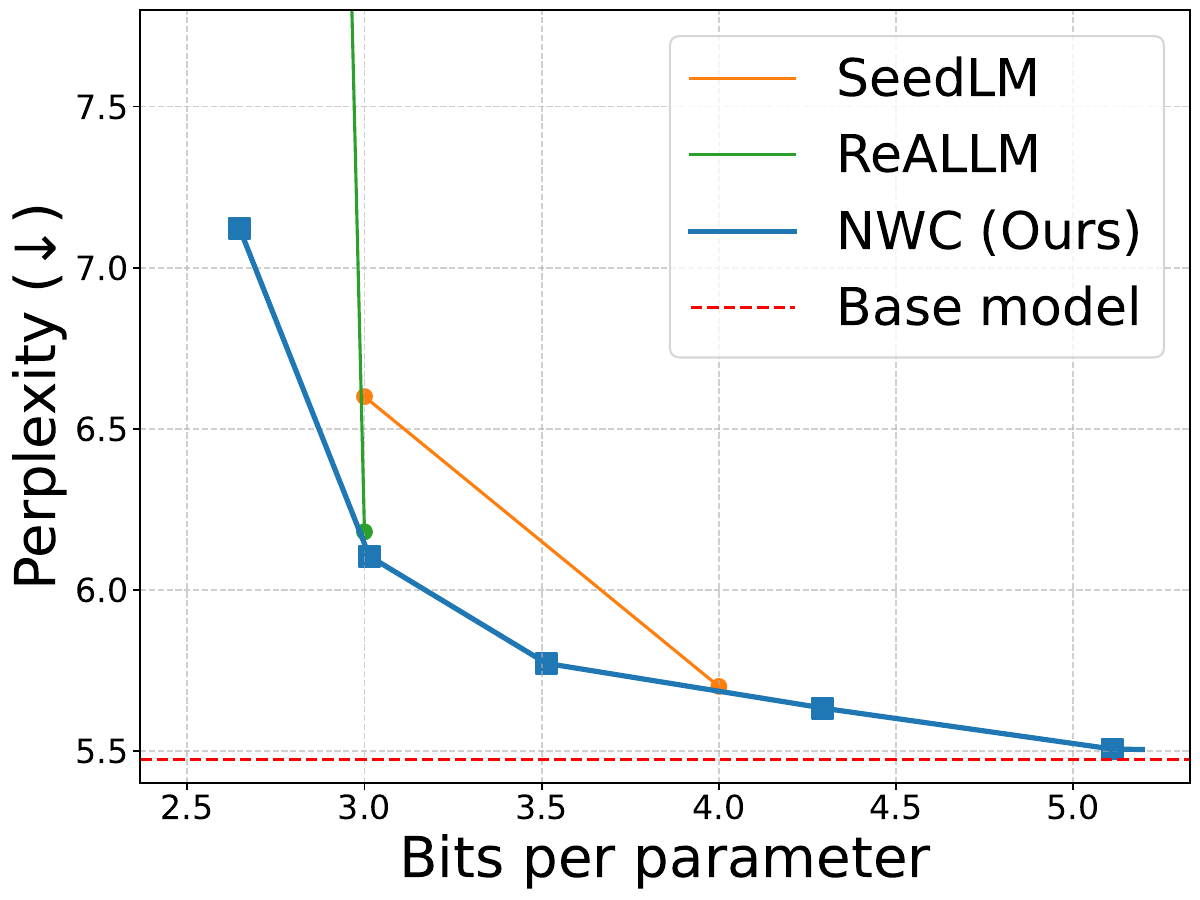}
        \caption{Llama 2-7B}
        \label{fig:7b_seedlm}
    \end{subfigure}
    \caption{Data-free compression comparison against SeedLM and ReALLM. WikiText-2 perplexity of Llama models compressed without calibration data.}
    \label{fig:seedlm}
    \vspace{-1.5em}
\end{figure}

%% file: sections/5.analyses.tex
\section{Analysis}

\subsection{On entropy-constrained quantization}\label{ssec:entropy_coding}

Why is NWC effective, especially at 4--6 bits? Our analysis suggests that this may be due to the effectiveness of the entropy-constrained quantization---i.e., jointly optimized rate and distortion---in handling heavy-tailed distributions at higher rates, where prior approach falls suboptimal.

To show this, we compare the rate-distortion curve of various compression schemes---scalar Lloyd-Max quantization (SQ), vector quantization (VQ), trellis-coded quantization (TCQ), and entropy-constrained scalar quantization (ECSQ)---across various data distributions; for TCQ, we adopt the configuration of QTIP \citep{qtip}. We also evaluate a variant of TCQ, termed TCQ$^*$, which uses the variable-sized codebook optimized for the rate. Furthermore, for the Laplace and model weight sources, the codebooks were initialized using a Laplace distribution instead of the standard normal distribution. Details regarding the experimental setup are provided in \Cref{sec:role_entropy_setup}.

\input{figure_tex/codec_rd}

In \Cref{fig:codec_rd}, we observe that ECSQ stays close to the Shannon limit across all sources and rates \citep{Gish1968AsymptoticallyEQ}.  In contrast, fixed-rate methods exhibit a gap that widens as the rate increases. Variants with relaxed codebook constraints, such as TCQ$^*$ and TCQ (RPTC), also fail to be near-optimal, especially in heavy-tailed distributions. While random Hadamard transform (RHT) can mitigate this, the gap persists at high rates. ECSQ yields superior and robust performance, particularly at $\ge 4$ bits.

\subsection{On learned transforms}

If ECSQ alone can achieve near-Shannon-limit MSE, what is the role of the learned transforms? Experiments suggest that the transform helps ensure a good model quality, which is not fully guaranteed by having low MSE.

In \Cref{fig:wiki_ecsq}, we find that while adding learned transforms to ECSQ slightly worsens the MSE, it can greatly reduce the perplexity of the compressed LLM. This observation suggests that the learned transform effectively steers the compression process to retain weight components essential for model performance, rather than na\"{i}ve MSE.

\input{figure_tex/Llama3-8B_ppl_wikitext2_ecsq}
\input{figure_tex/transform_qualtiy}

To understand which transform has been learned, we analyze the structural properties via the Jacobian $J \in \mathbb{R}^{d \times d}$, which characterizes the local linear behavior of a transform over the data distribution (see Appendix \ref{sec:empirical_jacobian} for detailed calculations). 
\input{tables/outlier_suppress}
We measure two properties:
(i)~\emph{Orthogonality}, $\|J^\top J - I\|_F / \|I\|_F$, which equals zero 
for a perfectly orthogonal transform; and
(ii)~\emph{Participation ratio}, $\mathrm{PR}(J) = \sum_j \|J_{:,j}\|_1^2 / \|J_{:,j}\|_2^2$, normalized to $[0,1]$, where $1$ corresponds to the maximally spread RHT and $0$ to a one-hot (identity-like) transform.
As shown in \Cref{fig:transform_metric}, the learned transform achieves near-orthogonality, yet its participation ratio remains below that of RHT, indicating that full rotation of every 
channel is not necessary. This finding is consistent with \citet{liang2025paroquant}, who similarly observe that selective rotation suffices for effective quantization.

Finally, we note that the learned transform also has an effect of suppressing outliers and heavy-tails (\Cref{tab:outlier_suppress}). Here, we observe significantly reduced kurtosis and outlier frequency for features yielded by neural encoders, rendering the representation more amenable to coding. 

\input{tables/latency}
\subsection{Latency analysis}\label{sec:latency}

In \Cref{tab:latency}, we assess the computational efficiency of NWC by measuring the wall-clock latency for compressing and decompressing a $4096 \times 4096$ weight. We have evaluated at the target rate of 4 bits per parameter; see \Cref{app:latency} for details.

From the table, we observe that the NWC achieves the latencies that are competitive to prior quantization methods, despite the prevalent belief that entropy-coding-based methods are slow. Indeed, it outperforms vector quantization baselines in terms of the encoding speed, and performs similarly in terms of the decoding speed. This is largely due to the availability of GPU-accelerated entropy decoding library---namely, the ``nvCOMP'' library by NVIDIA.\footnote{\url{https://developer.nvidia.com/nvcomp}\label{nvcomp}} These results highlight the proposed neural codec may have some potential to be further developed into a practical tool in the near future.

%% file: figure_tex/codec_rd.tex
\begin{figure}[t!]
    \centering
    \begin{subfigure}[b]{0.495\columnwidth}
        \centering
        \includegraphics[width=\columnwidth]{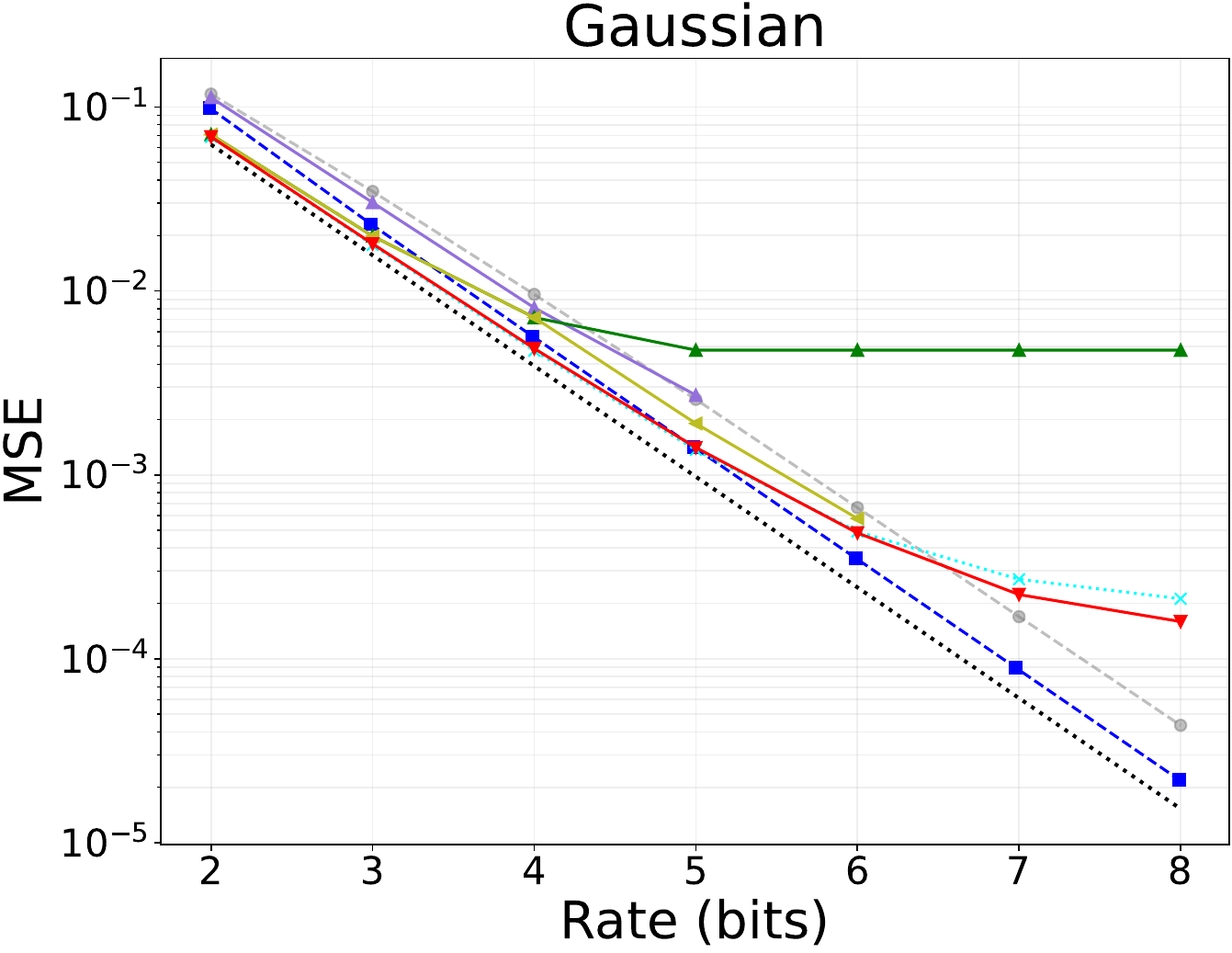}
        \label{fig:quant}
    \end{subfigure}%
    \hfill
    \begin{subfigure}[b]{0.495\columnwidth}
        \centering
        \includegraphics[width=\columnwidth]{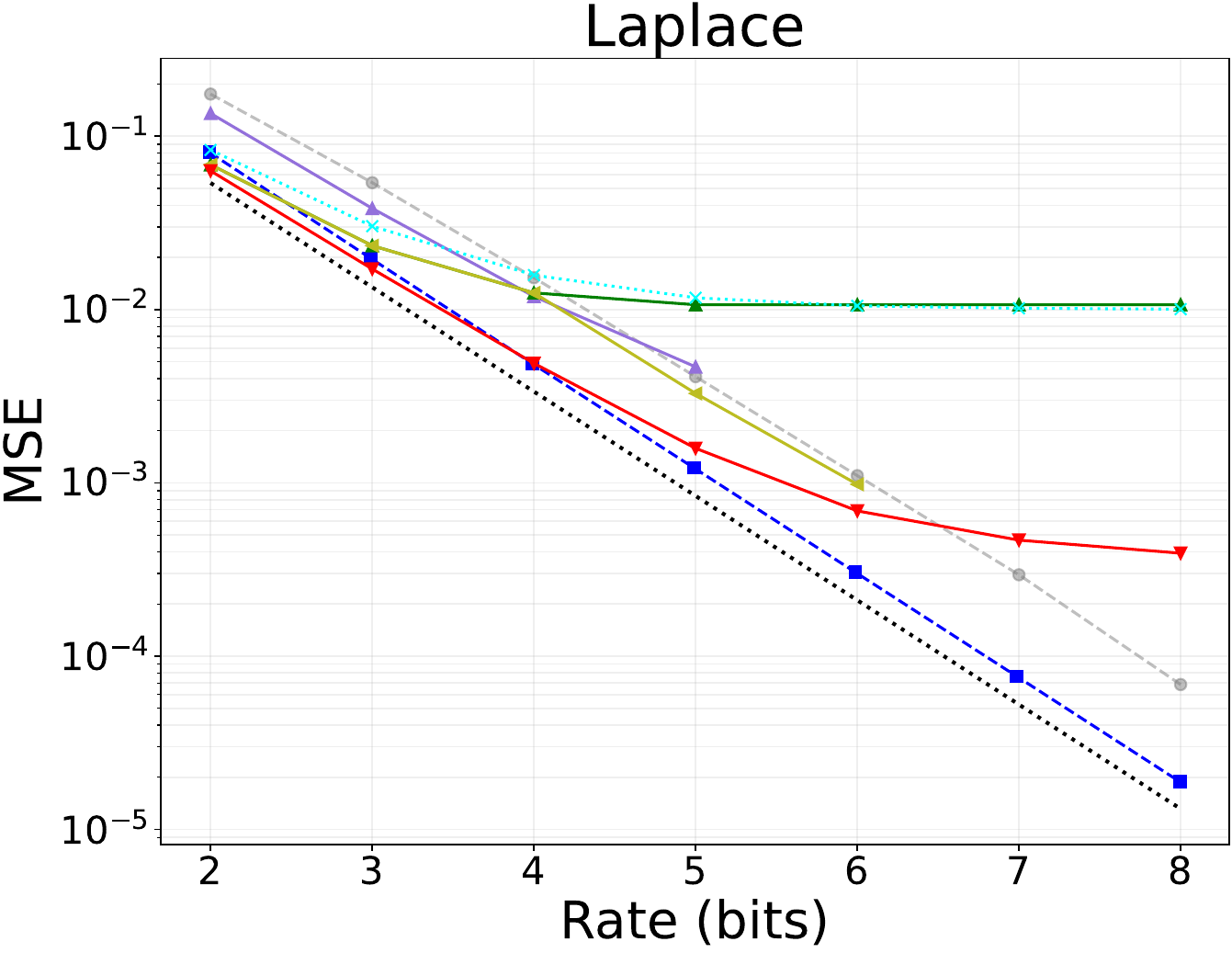}
        \label{fig:decomp}
    \end{subfigure}
    
    \vspace{-1em} 
    
    \begin{subfigure}[b]{0.495\columnwidth}
        \centering
        \includegraphics[width=\columnwidth]{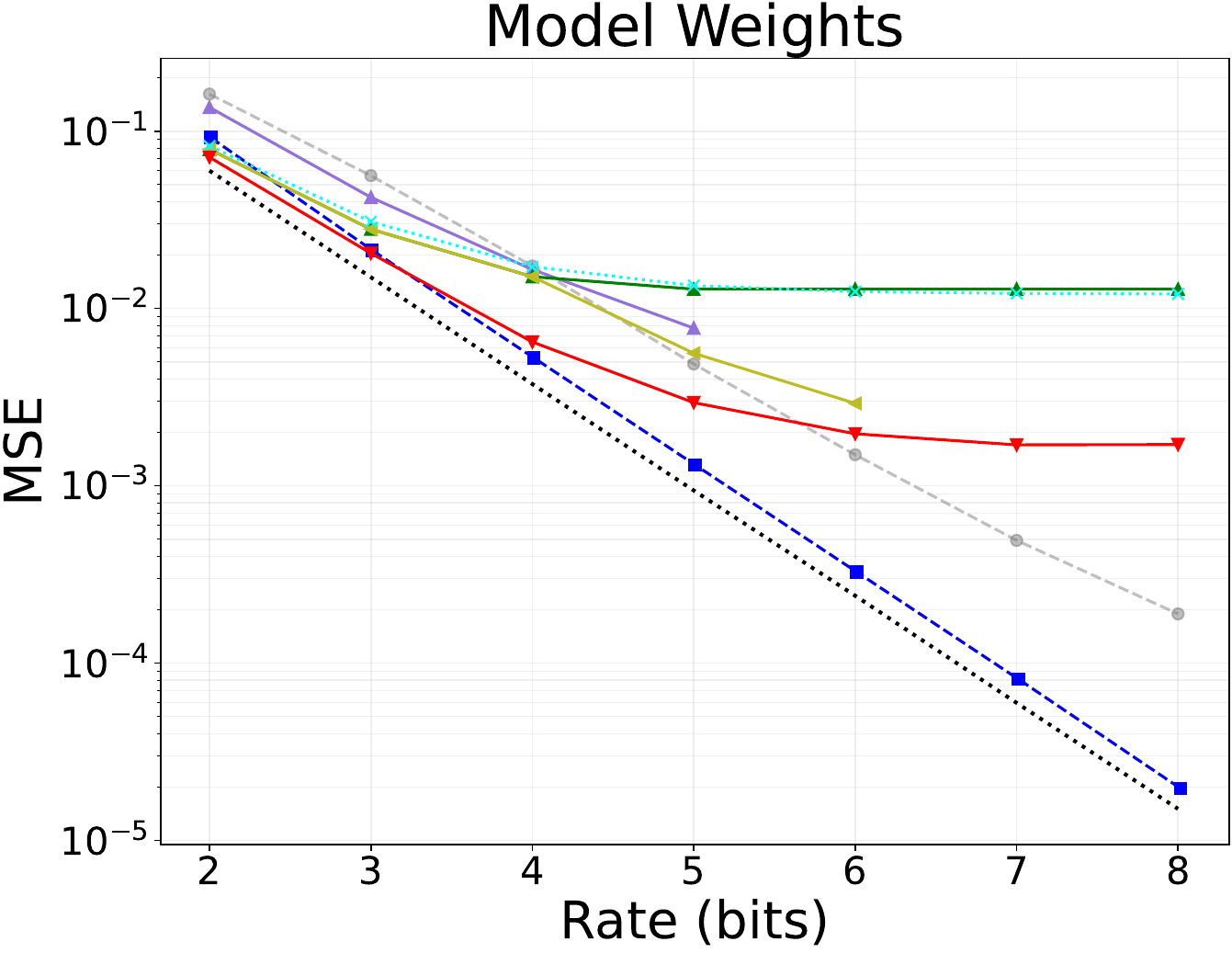}
        \label{fig:neural1}
    \end{subfigure}%
    \hfill
    \begin{subfigure}[b]{0.495\columnwidth}
        \centering
        \includegraphics[width=\columnwidth]{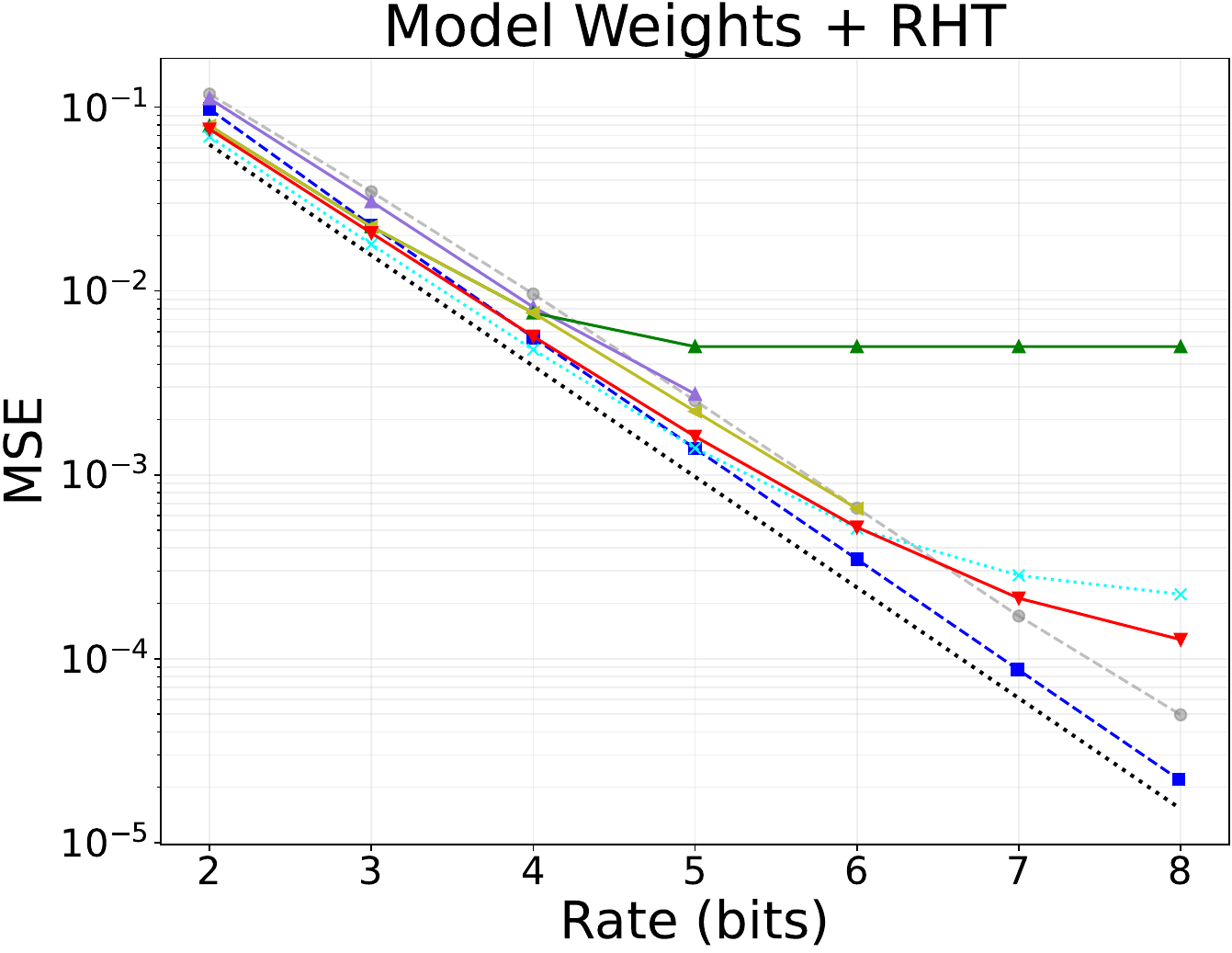}
        \label{fig:neural2}
    \end{subfigure}
    
    \vspace{-1em} 
    
    \begin{subfigure}[b]{0.85\columnwidth}
        \centering
        \includegraphics[width=\columnwidth]{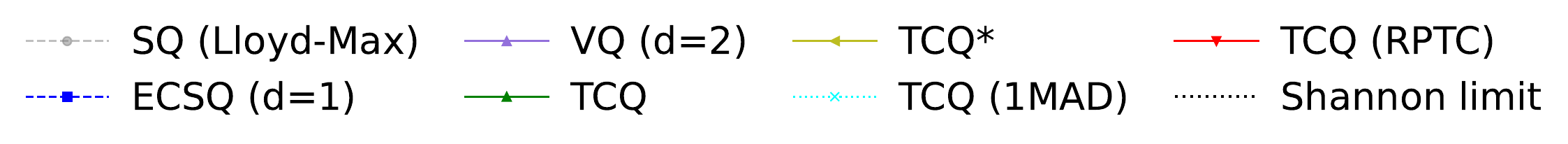}
    \end{subfigure}
    
    \caption{Rate-distortion curves for various quantization schemes and data distributions. The entropy constrained scalar quantization (ECSQ) remains close to the Shannon limit at high rates, while fixed-rate methods (SQ, VQ, TCQ, TCQ*) become suboptimal. RHT denotes the random Hadamard transform, and the model weights are from the Llama 3-8B.}
    \label{fig:codec_rd}
    \vspace{-1em}
\end{figure}

%% file: figure_tex/Llama3-8B_ppl_wikitext2_ecsq.tex
\begin{figure}[t!]
    \centering
    \begin{subfigure}[b]{0.48\columnwidth}
        \centering
        \includegraphics[width=\columnwidth]{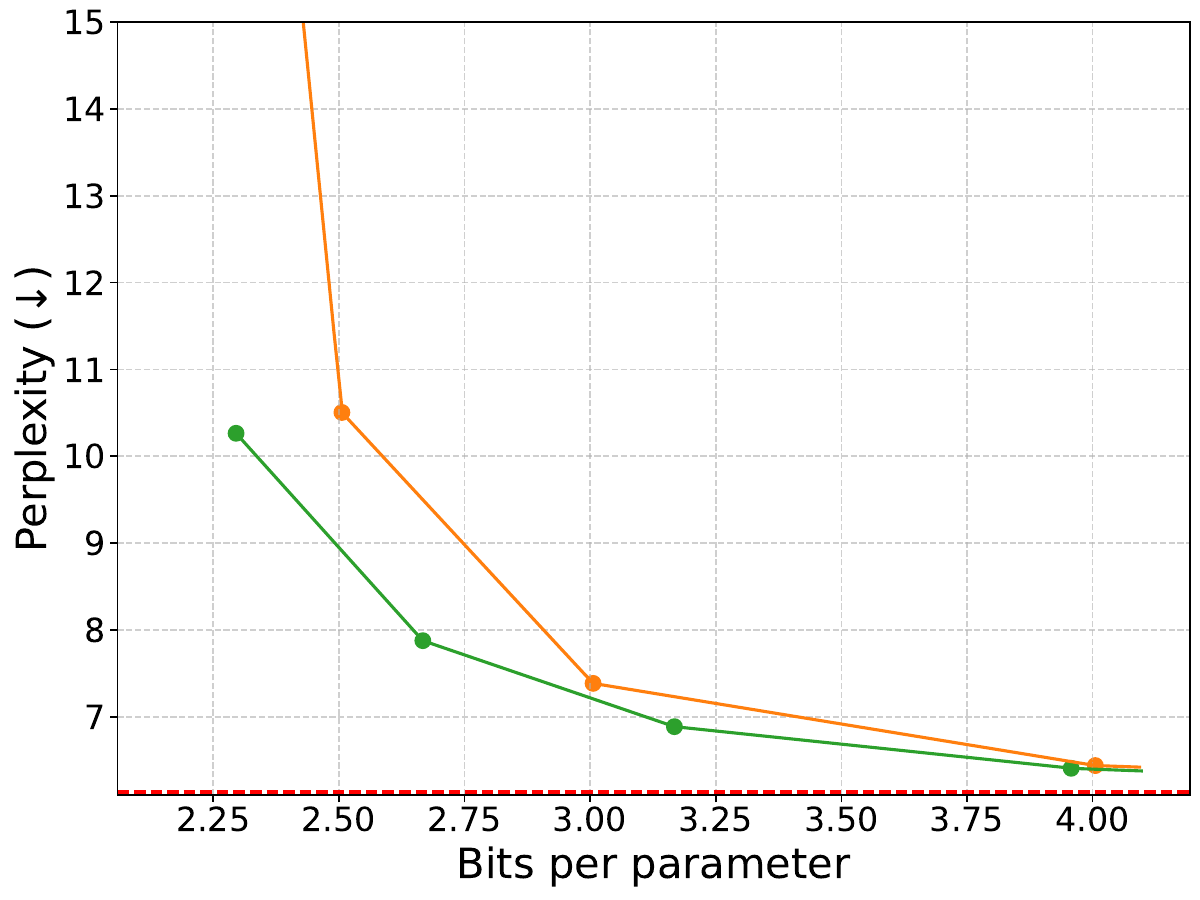}
        \caption{WikiText-2 perplexity}
    \end{subfigure}%
    \hfill
    \begin{subfigure}[b]{0.48\columnwidth}
        \centering
        \includegraphics[width=\columnwidth]{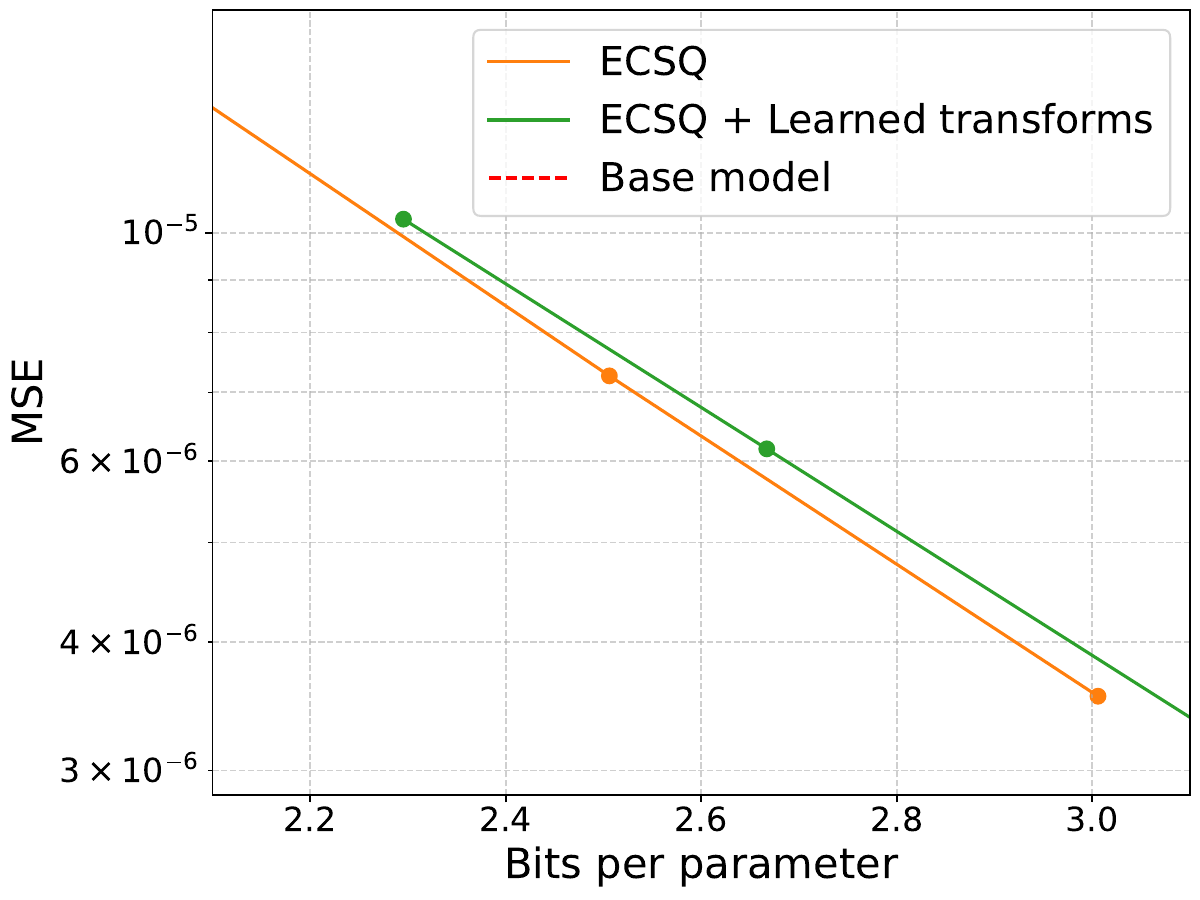}
        \caption{MSE}
    \end{subfigure}%
    \caption{Perplexity vs. MSE analysis on Llama 3-8B. The learned compression yields superior downstream performance despite higher reconstruction error.}
    \label{fig:wiki_ecsq}
    \vspace{-1em}
\end{figure}

%% file: figure_tex/transform_qualtiy.tex
\begin{figure}[t]
    \centering
    \includegraphics[width=1\columnwidth]{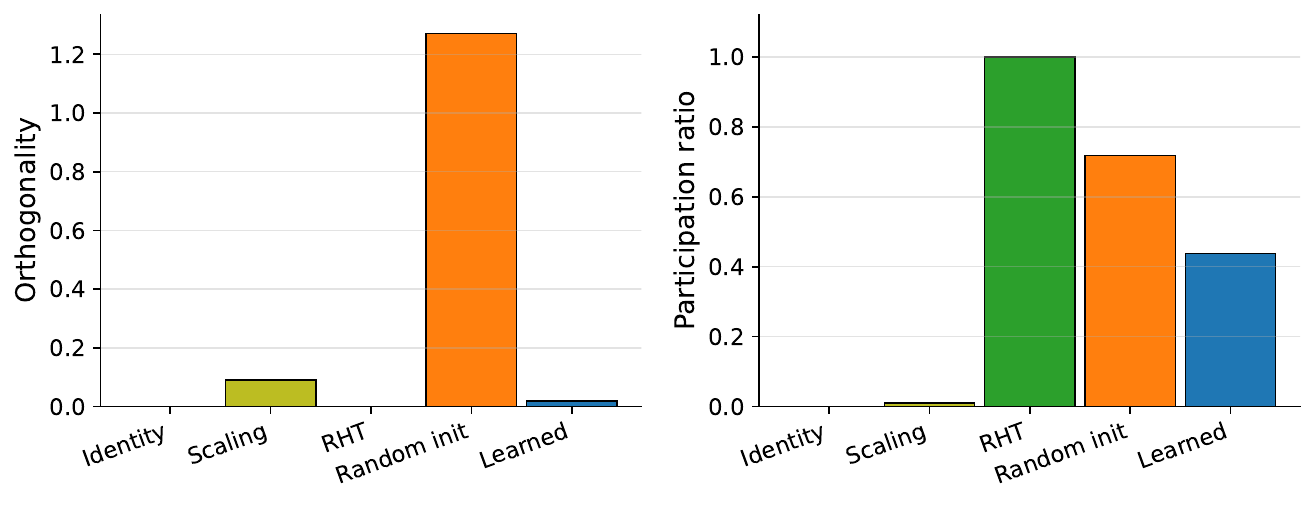}
    \caption{Comparison of transforms on two structural metrics measured over the first query projection (Llama 3-8B). (\textbf{Left}) Orthogonality: measures how close the empirical Jacobian is to an orthogonal matrix. (\textbf{Right}) Participation ratio: measures how uniformly the transform spreads energy across output dimensions, where 1 corresponds to the RHT.}
    \label{fig:transform_metric}
\end{figure}

%% file: tables/outlier_suppress.tex
\begin{table}[h!]
    \centering
    \resizebox{\columnwidth}{!}{
    \begin{tabular}{lccc}
        \toprule
        \textbf{Method} & \textbf{Kurtosis} & \textbf{Max Value ($\sigma$)} & \textbf{Outliers ($>3\sigma$)} \\
        \midrule
        Original        & 20.48 & 43.57 & 1.94\% \\
        DCT             & 0.46  & 10.54 & 0.50\% \\
        Random Rotation & 5.42  & 16.77 & 1.86\% \\
        RHT             & 5.44  & 16.66 & 1.86\% \\
        \textbf{NWC} & \textbf{0.00} & \textbf{2.70}  & \textbf{0.00\%} \\
        \bottomrule
    \end{tabular}
    }
    \caption{Statistical analysis of transformed query projection weights. The neural encoder significantly reduces kurtosis and outliers compared to other transformations.}\label{tab:outlier_suppress}
\end{table}

%% file: tables/latency.tex

\newcommand{\res}[2]{#1 \color{gray}{\scriptsize{$\pm$#2}}}

    

\begin{table}[t]
    \centering
        \resizebox{\columnwidth}{!}{
        \begin{tabular}{lcccc}
            \toprule
            \textbf{Operation} & \textbf{NWC (Ours)} & \textbf{GPTQ}  & \textbf{QTIP} & \textbf{QuIP\#} \\ 
            \midrule
            Encoding ($\times10^3$ ms) & \res{1.64}{0.30}  & \res{0.69}{0.02} & \res{19.84}{0.30} & \res{21.71}{0.82}  \\ 
            Decoding (ms)  & \res{1.17}{0.07}  & \res{0.08}{0.04}  & \res{0.52}{0.02} & \res{1.33}{0.33}  \\
            \hspace{1em} Synthesis, $g$ & \res{1.07}{0.07} & - & - & - \\
            \hspace{1em} Entropy Decoding & \res{0.10}{0.01} & - & - & - \\
            \bottomrule
        \end{tabular}
        }
    \caption{Encoding and decoding latency for $4096 \times 4096$ tensor, measured on a single NVIDIA RTX 6000 Ada.}\label{tab:latency}
    \vspace{-1em}
\end{table}

%% file: sections/6.related.tex
\section{Related work}\label{sec:related}

\textbf{Neural data compression.}
Unlike handcrafted codecs, neural codecs directly fit the data distribution and can optimize arbitrary differentiable distortion measures beyond MSE \citep{zhang2018unreasonable}. This flexibility has enabled strong performance on high-dimensional natural signals, including images \citep{balle, elic, ftic, tcm}, videos \citep{lu19,jia25}, and audio \citep{zeghidour21,defossez23high}. Inspired by these advances, we extend the neural compression paradigm to \textit{language model weights}, where the data are high-dimensional, heterogeneous, and must be preserved with respect to downstream model performance.

\noindent\textbf{Weight compression via neural codecs.}
Prior approaches to neural weight compression, such as autoencoders \citep{offlinecomp,vaecomp}, mainly target small-scale models such as BERT or LSTMs. Recent VQ-VAE-based methods have attempted to compress LLM weights \citep{reallm, tian2025pocketllm}, but they often rely on a \textit{piecemeal strategy}, such as overfitting codes or tuning decoders for individual weight matrices, similar to implicit neural representation codecs \citep{chen23}. In contrast, we train a single unified encoder-decoder pair to compress the entire set of LLM weights. To the best of our knowledge, this is the first \textit{global} neural compression framework that effectively scales to LLMs.

\noindent\textbf{Weight compression via quantization.}
Quantization is the dominant approach for LLM compression. Early methods focus on scalar quantization with second-order error compensation \citep{gptq} or activation-aware scaling \citep{awq}. Later works improve robustness to outliers using incoherent transformations, such as random Hadamard transforms \citep{quip, quarot}, while recent vector quantization methods push compression to sub-3-bit regimes \citep{aqlm, quip-sharp, qtip}. Although some methods use data-driven components such as gradient-based codebook optimization \citep{aqlm, cai2024lcq, qtip}, our framework differs by learning nonlinear transformations and entropy coding with a learnable probability model, rather than relying on handcrafted transforms.

\noindent\textbf{Other approaches.}
Other weight compression methods include pruning \citep{obs, pruning, simple}, low-rank approximation \citep{fwsvd, asvd, svdllm, Modegpt}, and seed-based representations \citep{seedlm}. These methods typically impose a very small decoding footprint, often allowing only one or no matrix multiplication. In contrast, our framework allows multi-layer decoders and entropy decoding, while keeping decoding latency competitive.

%% file: sections/7.conclusion.tex
\section{Conclusion}\label{sec:conclusion}
\vspace{-0.5em}
In this work, we introduce Neural Weight Compression, a novel framework that treats model weights as a learnable modality. We hope this work serves as a pioneering step toward fully automated compression pipelines in the era of large-scale models.

The NWC framework opens up the possibility of leveraging recent advances from the neural data compression literature. For instance, incorporating techniques such as progressive coding~\citep{progressive, progressive2, progressive3} and direct code optimization~\citep{codeeiting, overfitting} is a promising direction for future work. Exploring these avenues could further enhance both the performance and flexibility of the neural weight compression.

\section*{Limitations}
Our current implementation relies on tensor-level simulations, due to the nontrivial nature of implementing fused kernels for tile-wise arithmetic decoding and memory constraints of decoding parameters. We believe that bridging this gap through dedicated efforts will unlock meaningful inference acceleration as a promising future direction.

%% file: sections/99.appendix.tex
\clearpage
\appendix
\section*{\Large Appendix}
\section{Implementation Details}

\subsection{Experiment setup details}\label{ssec:setup_details}
For a thorough comparison, we reproduce the results of the baselines over the bitwidths missing in the original paper. For closed-source models, we compare with the figures in the original papers. Furthermore, to evaluate the efficacy of the compression scheme in isolation, we compare with the baselines without end-to-end fine-tuning. To clarify this point, we mark the modified baselines---QuIP\# and QTIP---with an asterisk ($^*$). For all experiments, we report the effective bits per parameter, which explicitly accounts for the overhead of metadata, including normalization factors and importance level indices.

\subsection{Reproduction of baselines over bitwidths} 
For QTIP, AWQ, and SpinQuant, we reproduce the results of the baselines over the bitwidths missing in original papers. 
\begin{itemize}[leftmargin=*,topsep=0pt,parsep=0pt]
    \item QTIP: We change $k$ in hybrid-computed codes
    \item SpinQuant: We optimize the W$b$A16KV16 rotation scheme for each target bitwidth $b$, then apply PTQ using this rotation in $b$-bits weight quantization setting.
    \item AWQ: We apply $b$-bit weight quantization using group size of 128, which induces roughly  0.25 extra bits per parameter. This overhead is included in the \textit{bits per parameter} reported on the X-axis of the comparison plots. (e.g. \Cref{fig:8b_wiki_mmlu_common}, \Cref{fig:llama_c4})
\end{itemize}
In contrast, we were unable to reproduce results for QuIP\# at arbitrary bit-widths. This is because it relies on fixed codebooks for each specific bitrate targeted in the original paper, which does not allow for flexible bitrate adjustment.


\subsection{Hessian generation}
We generate Hessian matrices following the protocol of QuIP\#~\citep{quip-sharp}, using the exact same matrices for all baseline comparisons (NWC, QuIP\#, QTIP). We use the RedPajama dataset~\citep{weber2024redpajama} for all language models: 6,144 sequences (length 2,048) for the Llama 2/3 families, and 1,024 sequences for Mixtral-8x7B, Qwen3-30B-A3B, Qwen3, and GPT-OSS-20B. For vision models, Hessians are estimated using 512 samples from the Conceptual Captions Dataset~\citep{sharma2018conceptual}.

\subsection{Network design and training}\label{sec:network_design}

\paragraph{Hyperparameter.} Details regarding the architecture and training of our compression network are provided in \Cref{tab:hyperparameters}. We employ an auxiliary loss to update the quantile parameters of the entropy model, which is trained separately from the main rate-distortion objective.
\input{tables/settings}

{\paragraph{Training cost.} We train the codec with the following compute resources and data:
\begin{itemize}[leftmargin=*,topsep=0pt,parsep=0pt]
    \item Compute cost: The codec is trained for 60 epochs, requiring 11.45 hours on a single NVIDIA A6000 Ada GPU.
    \item Dataset: The codec training does not require any external datasets or text corpora, utilizing only the model weights. For computational efficiency during random sampling, we aggregated 64 weight chunks into a single training sample. Consequently, for LLaMA-3-8B, the training set consists of approximately 6.8 million examples, with a validation set of 1,000 samples.
\end{itemize}


\subsection{Experiment details for vision models}\label{sec:vision_details}

For the vision encoders, we use the same codec pre-trained on the Llama weights from our LLM experiments. Note that no inter-layer recovery fine-tuning is applied to the either method, which we denote with $^\dagger$.

The specific layers targeted for compression in the vision encoders are as follows:
\begin{itemize}
    \item CLIP and SigLIP, compression was applied to all projection layers within both the text encoder and vision encoder blocks.
    \item DINOv2, compression was applied to all layers within the main encoder blocks.
\end{itemize}

\subsection{Block-wise inter-layer recovery fine-tuning for MoE models}\label{ssec:moe_finetuning}
Recent Mixture-of-Experts (MoE) architectures, such as Mixtral \citep{jiang2024mixtral}, and Qwen \citep{yang2025qwen3}, contain a significantly larger number of linear layers within a single Transformer block due to the multiplicity of experts. Consequently, performing block-wise recovery fine-tuning after compressing every single layer incurs prohibitive computational overhead. To address this we adopted a modified strategy that applies fine-tuning to groups of layers in the experiment of \Cref{fig:other_models} and \Cref{fig:other_models_common} for NWC and QTIP. Specifically, we execute the recovery fine-tuning following these groups of layers:
\begin{itemize}[leftmargin=*,topsep=0pt,parsep=0pt]
    \item Compress the self-attention weights ($\mathbf{W}_q, \mathbf{W}_k, \mathbf{W}_v, \mathbf{W}_o$) and the router weights, followed by block fine-tuning.
    \item Compress the up-projection weights of \textit{all} experts, followed by block fine-tuning.
    \item Compress the gate-projection weights of \textit{all} experts, followed by block fine-tuning.
    \item Compress the down-projection weights of \textit{all} experts, followed by block fine-tuning.
\end{itemize}


\subsection{Experiment details of \Cref{fig:codec_rd}}\label{sec:role_entropy_setup}

To substantiate the efficacy of entropy coding and optimization via the joint rate-distortion objective (\Cref{eq:learned_rd_objective}), we compare the rate-distortion performance of various compression schemes—scalar Lloyd-Max quantization (SQ), vector quantization (VQ), trellis coded quantization (TCQ), and entropy-constrained scalar quantization (ECSQ)—across i.i.d. Gaussian and Laplacian sources, and pre-trained language model weights. For all evaluations, we fix a sample size of $2^{20}$

For TCQ baselines, we evaluate several configurations to ensure a rigorous comparison:
\begin{itemize}[leftmargin=*,topsep=0pt,parsep=0pt]
    \item \textbf{Standard TCQ:} We adopt the default configuration from QTIP \citep{qtip} with constraint length $L=16$, trellis states $V=2$, and a fixed codebook size $Q=9$.
    \item \textbf{TCQ$^*$ (Scaled codebook):} To address performance saturation at higher bitrates (specifically $\ge 5$ bits), we employ a variant where the codebook size scales dynamically with the target rate ($Q = 2 \times \text{rate} + 1$). Furthermore, for Laplace and model weight sources, codebooks are initialized using a Laplace distribution rather than the standard normal distribution.
    \item \textbf{TCQ variants:} We also include the Random Permutation Trellis Coder (RPTC) as implemented in QTIP, and the Lookup-Free Computed Codebook (1MAD) variant to benchmark against diverse TCQ optimizations.
\end{itemize}

\subsection{Calculation of the Empirical Jacobian}
\label{sec:empirical_jacobian}
To analyze the structural properties of the learned transforms, we utilize the empirical Jacobian $J \in \mathbb{R}^{d \times d}$, which captures the local linear behavior of a transform $T$. The matrix $J$ is computed as:
$$J = \Sigma_{zx}\Sigma_{xx}^{-1}$$
where $\Sigma_{zx}$ and $\Sigma_{xx}$ are the cross-covariance and auto-covariance matrices of the output $z = T(x)$ and the input $x$, respectively.

\subsection{Details of latency analysis}\label{app:latency}
In this section, we provide details of the latency analysis presented in \Cref{sec:latency}. The reported values represent the average wall-clock time and standard deviation for processing a $4096 \times 4096$ weight tensor at a target rate of 4 bits per parameter on NVIDIA RTX A6000.
\begin{itemize}[leftmargin=*,topsep=0pt,parsep=0pt]
    \item \textbf{Encoding Time:} This measures the total duration to transform the original tensor into its compressed format. Crucially, this includes the computational overhead of the Hessian-based intra-layer error feedback mechanism (e.g., LDLQ). Note that we exclude the runtime for inter-layer recovery fine-tuning for NWC, QuIP\#, and QTIP. This is because this process operates at the block level---making tensor-wise attribution ambiguous---and incurs a comparable computational cost across all methods.
    \item \textbf{Decoding Time:} This corresponds to the latency involved in fully reconstructing the full precision tensor in VRAM from the compressed bitstream. For NWC, the total decoding time is calculated as the sum of the neural synthesis latency and the entropy decoding time. Specifically, the entropy decoding component is benchmarked separately using the Asymmetric Numeral Systems (ANS) codec from the NVIDIA nvCOMP library\textsuperscript{\ref{nvcomp}}.
\end{itemize}



\input{figure_tex/llama_c4}
\input{figure_tex/result_other_models_common}

\section{Additional Results}\label{sec:additional}

\subsection{Language models}\label{ssec:additional_llm}

\paragraph{C4 perplexity.} \Cref{fig:llama_c4} presents the C4 perplexity of Llama3-8B, Llama2-7B, and Llama2-13B across various bit range.

\paragraph{Common-sense.} \Cref{fig:other_models_common} reports the average accuracy on 6 common-sense reasoning task.



\subsection{Comparison with other baselines}\label{ssec:nonptq}
{In the main experiments, we leave out comparing NWC with the QAT methods The main reason is that QAT methods often require significantly more computation than ours. Specifically, LLM-QAT demands approximately 280 hours on an A100 80G. Furthermore, QAT relies on the full pre-training corpus or extensive synthetic data, NWC requires only the small calibration set.}

\input{figures/Llama2-7B_QAT}
\paragraph{QAT \& ReALLM.} 
In this section, we compare NWC against Quantization-Aware Training: LLM-QAT~\citep{liu2023llm}, BitDistiller~\citep{du2024bitdistiller} and other neural codecs: ReALLM~\citep{reallm}. The evaluation is performed on Llama2-7B, measuring perplexity on WikiText-2 and the average zero-shot accuracy across four common-sense tasks (PiQA~\citep{piqa_dataset}, HellaSwag~\citep{hellaswag}, WinoGrande~\citep{winogrande_dataset}, and ARC-Challenge~\citep{arc_dataset}). As shown in \Cref{fig:comparision_qat}, NWC consistently demonstrates superior performance compared to QAT-based methods at equivalent bitrates. 
Crucially, when evaluated in a calibration data-free setting, NWC compression performance is substantially better.

\input{tables/comparison_svd}
\paragraph{SVD-based methods.} \Cref{tab:comprasison_svd} presents a comparison between NWC and SVD-based compression methods \cite{svdllm, wang2025svd}. We evaluate both the compression ratio and the zero-shot accuracy on a range of tasks, including OpenbookQA~\citep{mihaylov2018suit} and MathQA~\citep{amini2019mathqa}. Here, the compression ratio is defined ratio of total storage cost, calculated as the total number of bits in the compressed model divided by that of the original model. The results indicate that NWC achieves superior performance compared to SVD even at higher compression ratios.

\input{tables/pocktllm}
\paragraph{PocketLLM.} \Cref{tab:pocketllm} presents the perplexity comparison with PocketLLM \citep{tian2025pocketllm} on the WikiText-2 and C4 benchmarks with context length of 4096. Note that we do not include zero-shot results from the paper in this section. A direct comparison was not feasible because their reported baseline performance for the uncompressed model and other methods diverges from ours, presumably due to differences in the evaluation setup.


\input{figures/ablation_main}
\section{Ablation studies}\label{sec:ablation}
We systematically evaluate the efficacy of each technical component of the proposed framework. First, we observe that applying normalization at a finer granularity (i.e., channel-wise) yields the most favorable rate-distortion trade-off compared to coarser alternatives (\Cref{fig:abl_normalization}). Second, the analysis confirms that leveraging importance-awareness is crucial for minimizing perplexity, particularly in low bitrate regimes (\Cref{fig:abl_importance}). Furthermore, we validate the necessity of the inference-time error mitigation strategies; the results demonstrate that both intra- and inter-layer compensation mechanisms are beneficial for effectively recovering model performance during compression (\Cref{fig:abl_eror}). Finally, comparisons against fixed-rate baselines reveal the clear superiority of our learned entropy model over standard scalar and vector quantization approaches, as shown in \Cref{fig:abl_entropy}.

\subsection{Encoder and decoder networks}
To quantify the benefit of the learnable $f$ and $g$ networks, we compare the model against an entropy-only baseline. This baseline replaces $f$ and $g$ with identity functions and is trained using only the auxiliary and rate losses; without a distortion loss, it is incapable of controlling the bitrate or adapting reconstruction quality.
For this ablation, both importance awareness and error compensation were disabled. The results in \Cref{tab:ablation_encoder} demonstrate that the learnable networks offer a large reduction in perplexity compared to the entropy-only model.
\input{tables/ablation_encoder}

\subsection{Use of entropy model}
In \Cref{fig:abl_entropy}, to validate the effectiveness of density estimation-based entropy coding within our framework, we conducted an ablation study comparing our method against baselines where the entropy model is replaced by non-linear fixed-length quantizers. We trained and compared compression models utilizing a non-linear scalar quantizer, a vector quantizer, and the entropy model, respectively. For all three models, the encoder and decoder architectures were identical, and no input importance was employed. The two fixed-length quantizers were trained using the Straight-Through Estimator (STE), with the dimension of the vector quantizer set to 2.


\subsection{Importance aware compression}\label{sec:ablation_ql}
\Cref{fig:abl_importance} presents a comparison of compression performance between our importance-guided quality allocation and a baseline using a uniform quality assignment. The results demonstrate that leveraging importance consistently leads to lower perplexity. This performance gain is particularly significant in the low bitrate regime.

\subsection{Channel-wise normalization}
\Cref{fig:abl_normalization} presents an ablation study on the granularity of weight normalization. We compare the performance when normalization is applied at different levels: globally (across the entire model weights), tensor-wise (per-layer), and channel-wise. The results, which account for the bit overhead required to store the normalization factors, show a clear trend: a finer granularity (i.e., channel-wise) consistently leads to lower perplexity.

\subsection{Compensation for compression error}
The effectiveness of our inference-time compression error compensation is demonstrated in \Cref{fig:abl_eror}. The results show that applying each of the intra- and inter-layer error compensation mechanisms leads to a lower perplexity, confirming the positive contribution of both components. 

\input{tables/ablation_chunksize}
\subsection{Chunk size}
In \Cref{tab:ablation_chunksize}, we present an ablation study of weight chunk size. In general, NWC is robust against varying chunk sizes. We chose 16, which achieved the best result at low rates. However, one can use smaller chunk sizes to reduce the decoder size (although they are already small).

\input{figure_tex/ablation_K}
\subsection{Number of importance levels}
In \Cref{fig:ablation_K}, we present an ablation study of the number of importance level, $K$. The performance saturates quite early, at K=4. That is, 2bits are sufficient for the metadata.

\input{tables/ablation_hessian_assign}
\subsection{Hessian importance assignment}
In \Cref{tab:ablation_hess}, we compare the Hessian-based importance assignment with random assignment. This confirms that the Hessian-based assignment is indeed essential
 
\section{Additional Analyses}




\input{figures/layer_stats}
\input{figure_tex/weight_channel_box}

\subsection{Per-layer statistic of large language model}\label{ssec:per_layer_scale}
In \Cref{fig:layer_stats}, we visualize the kurtosis and standard deviation across different layer depths and types. From these results, we observe three key characteristics:
\begin{itemize}[leftmargin=*,topsep=0pt,parsep=0pt]
\item The majority of layers exhibit kurtosis values higher than that of a Gaussian distribution (i.e., $>0$), indicating heavy-tailed distribution.
\item Certain layers exhibit extremely high kurtosis. This is particularly pronounced in the $Q$, $K$ and $O$ projections of the first block.
\item The standard deviation varies depending on the layer type.
\end{itemize}

\subsection{Channel-wise scale}\label{ssec:channel_scale}
Similar to the observation in CNNs \citep{nagel2019data}, the transformer-based models exhibit scaling differences across channels. We visualize the channel-wise scale for the $Q$ matrix of Llama 3-8B in \Cref{fig:channel_box}.

\input{figures/layerwise_RD}
\subsection{Robustness to heavy-tailed distributions}
\Cref{fig:layerwise_rd} presents the single-layer rate-distortion curves for QTIP and NWC. Our method outperforms QTIP in terms of MSE, particularly in layers characterized by high kurtosis (i.e., heavy-tailed distributions). This observation aligns with \Cref{fig:codec_rd} and \Cref{tab:outlier_suppress}, demonstrating that learned codecs are more robust to extreme distributions and surpass handcrafted approaches in handling heavy-tailed weights.



\section{Details of Determination and Assignment of Importance Levels}\label{sec:importance_detail}

\textbf{Assigning importance levels during inference}. We determine the importance level for each column based on the magnitude of its corresponding Hessian diagonal elements. Specifically, we employ a quantile-based stratification strategy with thresholds that are fixed across all weight tensors. Columns in the top 0.1\% of Hessian values are assigned importance level 3, those in the top 1\% receive level 2, and those in the top 10\% are set to level 1. All remaining columns are assigned level 0. This ensures that bit allocation is prioritized for the most sensitive parameters.

\textbf{Importance level training coefficient}. During training, an importance level coefficient is sampled from a discrete set for each weight chunk. These values were empirically selected to span a wide spectrum of operating points, ensuring the model learns to handle diverse rate-distortion trade-offs. Following the literature on variable-rate neural image compression \citep{control_nic}, we formulate the loss by applying the reciprocal of the importance coefficient $\lambda_I$ to the rate term (i.e., weighting the rate by $1/\lambda_I$). We observe that this formulation enhances the model's ability to learn diverse bitrate controls compared to distortion weighting schemes.

\subsection{Comparison with AWQ and GPTQ}\label{sec:importance_comparison}

In this section, we explicitly compare the proposed NWC with existing post-training quantization methods like AWQ and GPTQ. Our method differs in two critical aspects:

\begin{itemize}[leftmargin=*,topsep=0pt,parsep=0pt]
    \item \textbf{Discrete importance:} For compact storage, we discretize the scaling factors into a small set of levels (e.g., $K=4$). These levels can be stored using only trivial overhead (e.g., 2 bits per channel). In contrast, the sensitivity metrics in GPTQ and AWQ are continuous.
    \item \textbf{Importance-augmentation during training:} To address the imbalance in samples across importance levels, we conduct importance-augmented training where each vector is paired with a randomly selected scaling factor.
\end{itemize}



\subsection{Activation-aware scaling in AWQ}
AWQ's activation-aware scaling is defined as 
\begin{align}
    \mathbf{s} = \mathrm{\mathbf{s_X}}^{\alpha^*},  \quad \alpha^* = \arg\min_{\alpha \in [0, 1]} \mathcal{L}(\mathbf{s_\mathbf{X}^\alpha})\
\end{align}
where  $(\mathbf{s_X})_c = \frac{1}{N} \sum_{i=1}^{N} |\mathbf{X}_{c,i}|$, and $\mathcal{L}(\mathbf{s})$ is the quantization objective of Equation (4) in \citet{awq}.

This scaling mechanism relies on the empirical assumption that the quantization step size $\Delta = \max(|\mathbf{w}|)$, remains stable even after scaling (i.e., $\Delta' = \max(|\mathbf{s^\top w|}) \approx \Delta$).
As the scaling factor $s$ increases, this assumption breaks down, forcing AWQ to use a heuristic search to find a safe $\alpha$.

As a fully learnable framework, NWC can explicitly incorporate the scale into its loss function, allowing the optimizer to find the most effective compression directly with respect to the importance. 

\subsection{Sensitivity metrics in GPTQ}
Drawing from OBQ~\citep{obq}, GPTQ leverages arbitrary ordering and Cholesky reformulation to define the error sensitivity metrics of each weight column as: 
\begin{align}
    \mathbf{s} = \mathrm{diag}(\mathbf{L}^\top)^{-1}, \quad \mathbf{L} = \mathrm{Cholesky}(\mathbf{H}^{-1}).
\end{align}

Adopting this as an importance metric, however, is computationally more intensive than simply using the Hessian diagonal $\mathbf{\Lambda}$. It requires one additional matrix inversion and one Cholesky decomposition for each Hessian matrix.
Empirically, we find that using the diagonal of the Hessian directly is not only sufficient but often yields superior results for training neural compression models.

\section{{Justification of diagonal approximation}}
In \Cref{ssec:training}, we employ a diagonal approximation of the Hessian matrix to measure parameter sensitivity. While the full Hessian captures cross-parameter correlations, we justify the adequacy of the diagonal approximation based on the distinct activation characteristics--outlier activations-- in LLMs. 

LLMs are known to exhibit massive outlier activations, where specific feature dimensions possess magnitudes significantly larger than others \cite{sun2024massive, an2025systematic}. Let $\mathbf{x} \in \mathbb{R}^{d_{in}}$ denote the input activation. The Hessian is typically approximated using the expected outer product of the inputs, $H \approx \mathbb{E}[\mathbf{x}\mathbf{x}^\top]$.

Consider a feature dimension $d$ that corresponds to an outlier feature. Mathematically, if the magnitude of this outlier feature is significantly larger than other dimensions (i.e., $|x_d| \gg |x_j|$ for $j \neq d$), the diagonal term $H_{dd}$ dominates the off-diagonal terms:
\begin{align}
    H_{dd} \approx \mathbb{E}[x_d^2] \gg \mathbb{E}[x_d x_j] \approx H_{dj}    
\end{align}
The Hessian matrix becomes effectively diagonally dominant in the presence of strong outliers. Therefore, the diagonal approximation provides a simple and computationally efficient proxy for parameter sensitivity, especially when sensitivity is dominated by a small number of high-energy activation dimensions.

\input{tables/8b_all_results}

\section{Numerical results for Rate--Accuracy curves}
\label{app:numerical_results}

\Cref{tab:nwc_results,tab:nwc_mmlu,tab:vision_results} report the numerical values used to plot the rate--accuracy curves in \Cref{fig:8b_wiki_mmlu_common,fig:other_models,fig:vision_models}.

\section{The Use of Large Language Models}
We utilized a large language model (LLM) to refine the language and improve clarity in several sections of this paper. However, its use was strictly limited to improving the writing style; the LLM did not contribute to the research ideation or the core scientific content.

\section{Licensing of Pretrained Models and Datasets}
Our use of all pretrained models and benchmark datasets complies with their original licenses and is consistent with their intended use for academic research and evaluation. Detailed licensing information is provided in \Cref{tab:model-dataset-licenses}.


\begin{table*}[t]
\centering
\small
\resizebox{\textwidth}{!}{%
\begin{tabular}{lll}
\toprule
\textbf{Model/Dataset} & \textbf{URL} & \textbf{License / Terms} \\
\midrule
Meta-Llama-3-8B & \url{https://huggingface.co/meta-llama/Meta-Llama-3-8B} & Meta Llama 3 Community License \\
Llama-2-7b-hf & \url{https://huggingface.co/meta-llama/Llama-2-7b-hf} & Llama 2 Community License \\
Llama-2-13b-hf & \url{https://huggingface.co/meta-llama/Llama-2-13b-hf} & Llama 2 Community License \\
Qwen3-30B-A3B & \url{https://huggingface.co/Qwen/Qwen3-30B-A3B} & Apache-2.0 \\
Qwen3-8B & \url{https://huggingface.co/Qwen/Qwen3-8B} & Apache-2.0 \\
Qwen3-4B & \url{https://huggingface.co/Qwen/Qwen3-4B} & Apache-2.0 \\
Mixtral-8x7B-v0.1 & \url{https://huggingface.co/mistralai/Mixtral-8x7B-v0.1} & Apache-2.0 \\
GPT-OSS-20B & \url{https://huggingface.co/openai/gpt-oss-20b} & Apache-2.0 \\
CLIP-ViT-L & \url{https://huggingface.co/openai/clip-vit-large-patch14} & MIT \\
SigLIP-B/16 & \url{https://huggingface.co/google/siglip-base-patch16-224} & Apache-2.0 \\
DINOv2-L & \url{https://huggingface.co/facebook/dinov2-large} & Apache-2.0 \\
\midrule
RedPajama & \url{https://huggingface.co/datasets/togethercomputer/RedPajama-Data-1T} & Per-subset licenses \\
Conceptual Captions & \url{https://huggingface.co/datasets/google-research-datasets/conceptual_captions} & Other / Conceptual Captions license \\
WikiText-2 & \url{https://huggingface.co/datasets/Salesforce/wikitext} & CC-BY-SA-3.0, GFDL \\
C4 & \url{https://huggingface.co/datasets/allenai/c4} & ODC-BY \\
MMLU & \url{https://huggingface.co/datasets/cais/mmlu} & MIT \\
ARC-Easy / ARC-Challenge & \url{https://huggingface.co/datasets/allenai/ai2_arc} & CC-BY-SA-4.0 \\
WinoGrande & \url{https://huggingface.co/datasets/allenai/winogrande} & Apache-2.0 \\
PIQA & \url{https://huggingface.co/datasets/ybisk/piqa} & Apache-2.0 \\
HellaSwag & \url{https://huggingface.co/datasets/Rowan/hellaswag} & MIT \\
BoolQ & \url{https://huggingface.co/datasets/google/boolq} & CC-BY-SA-3.0 \\
MMLU-Pro & \url{https://huggingface.co/datasets/TIGER-Lab/MMLU-Pro} & MIT \\
GPQA Diamond & \url{https://huggingface.co/datasets/Idavidrein/gpqa} & CC-BY-4.0 \\
AIME 2024 & \url{https://huggingface.co/datasets/HuggingFaceH4/aime_2024} & Apache-2.0 \\
AIME 2025 & \url{https://huggingface.co/datasets/math-ai/aime25} & Apache-2.0 \\
ImageNet-1k & \url{https://www.image-net.org/download.php} & ImageNet Terms of Access \\
\bottomrule
\end{tabular}%
}
\caption{Pretrained models and datasets used in our experiments.}
\label{tab:model-dataset-licenses}
\end{table*}

%% file: tables/settings.tex
\begin{table}[h]
\centering
\resizebox{\columnwidth}{!}{
\begin{tabular}{ll}
\toprule
\textbf{Hyperparameter} & \textbf{Value} \\ 
\midrule
Chunk size & 16 \\
$g_a$ network width & 512 \\
$g_a$ number of residual blocks & 4 \\
$g_s$ network width & 512 \\
$g_s$ number of residual blocks & 4 \\
Entropy model channel size & 16 \\
Learning rate & $1 \times 10^{-4}$ \\
Learning rate (auxiliary loss) & $1 \times 10^{-3}$ \\
Optimizer & Adam \\
$\lambda$ & $\{ 30, 50, 100, 300, 1000, 10000\}$ \\ 
$\lambda_I$ & $\{0.29,\ 0.83,\ 10,\ 20\}$ \\
\bottomrule
\end{tabular}
}
\caption{Hyperparameters for network design and training}\label{tab:hyperparameters}
\end{table}

%% file: figure_tex/llama_c4.tex
\begin{figure*}[t]
    \centering
    \begin{subfigure}[b]{0.325\textwidth}
        \includegraphics[width=\textwidth]{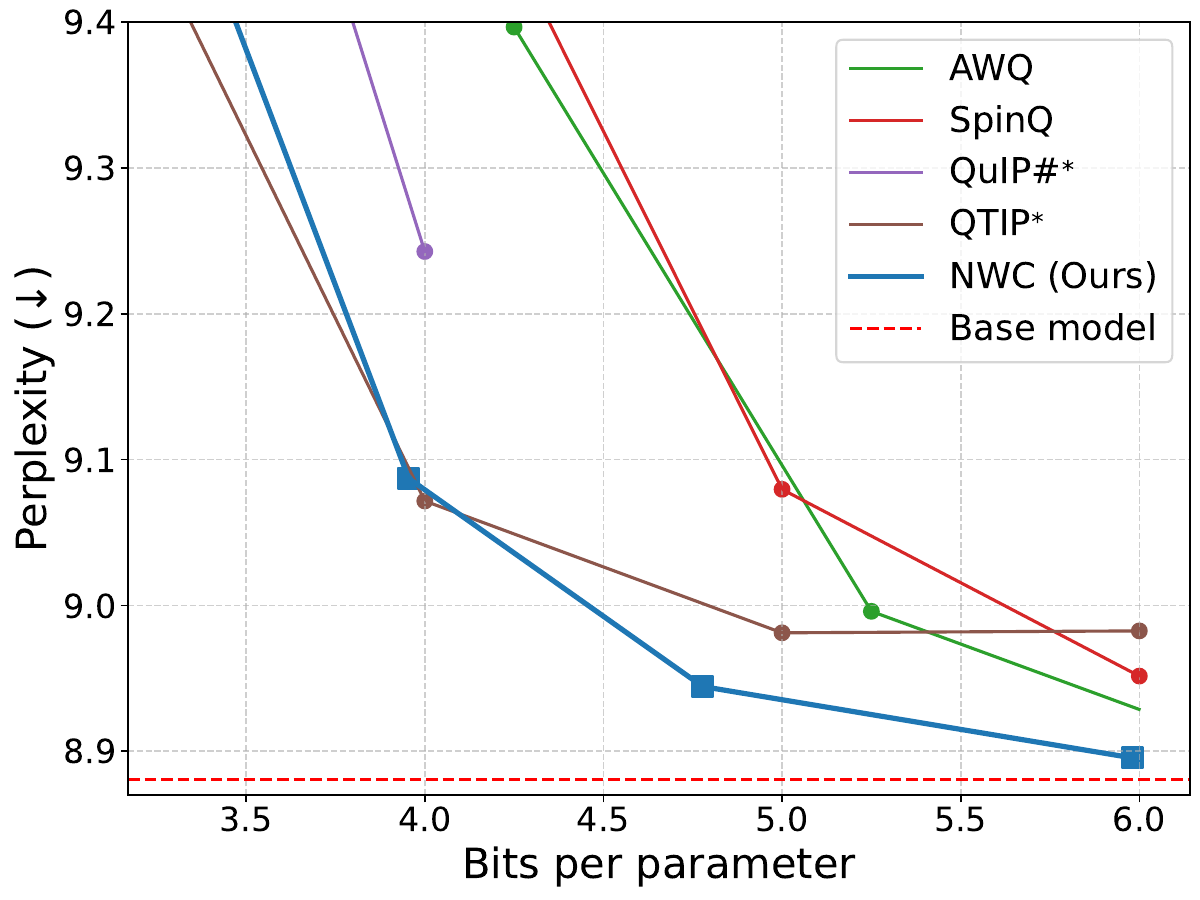}
        \caption{Llama 3-8B}
    \end{subfigure}
    \begin{subfigure}[b]{0.325\textwidth}
        \includegraphics[width=\textwidth]{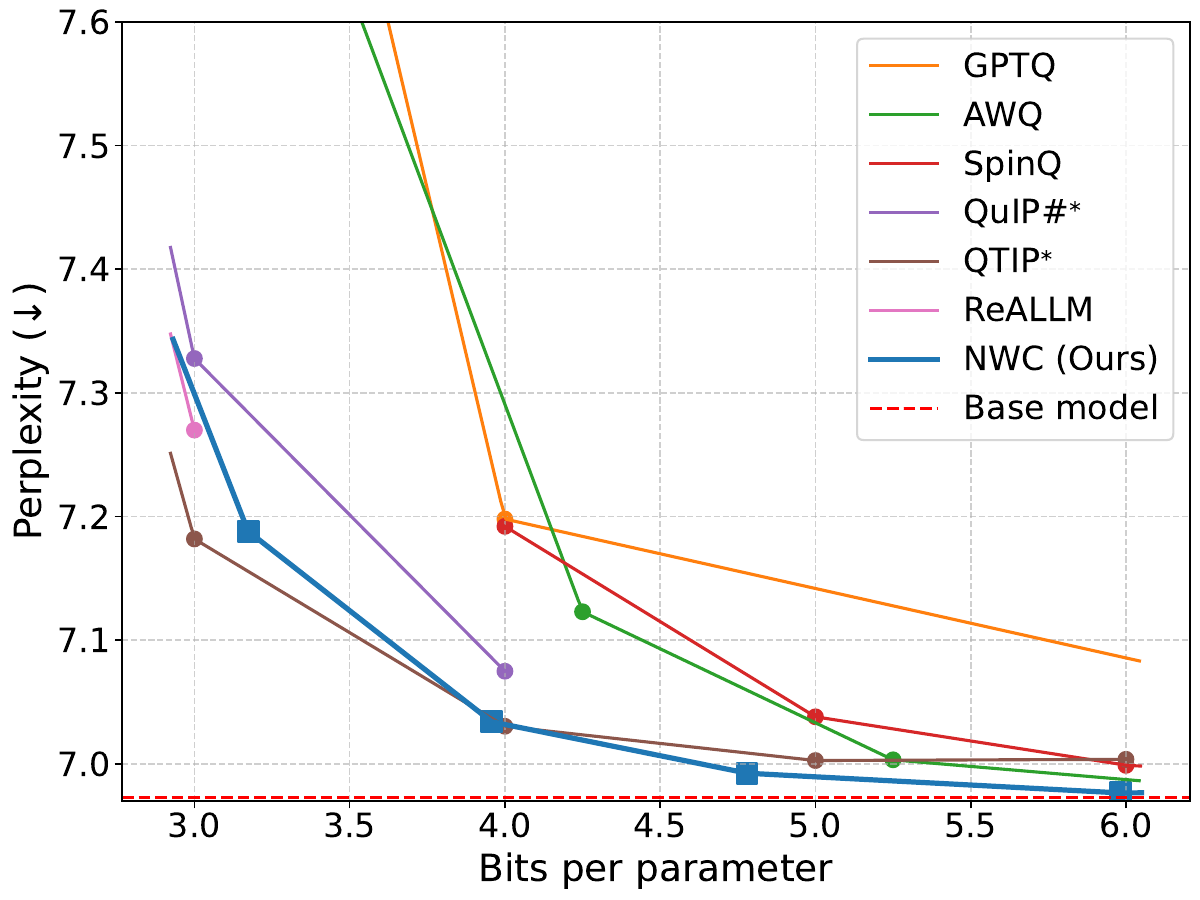}
        \caption{Llama 2-7B}
    \end{subfigure}
    \begin{subfigure}[b]{0.325\textwidth}
        \includegraphics[width=\textwidth]{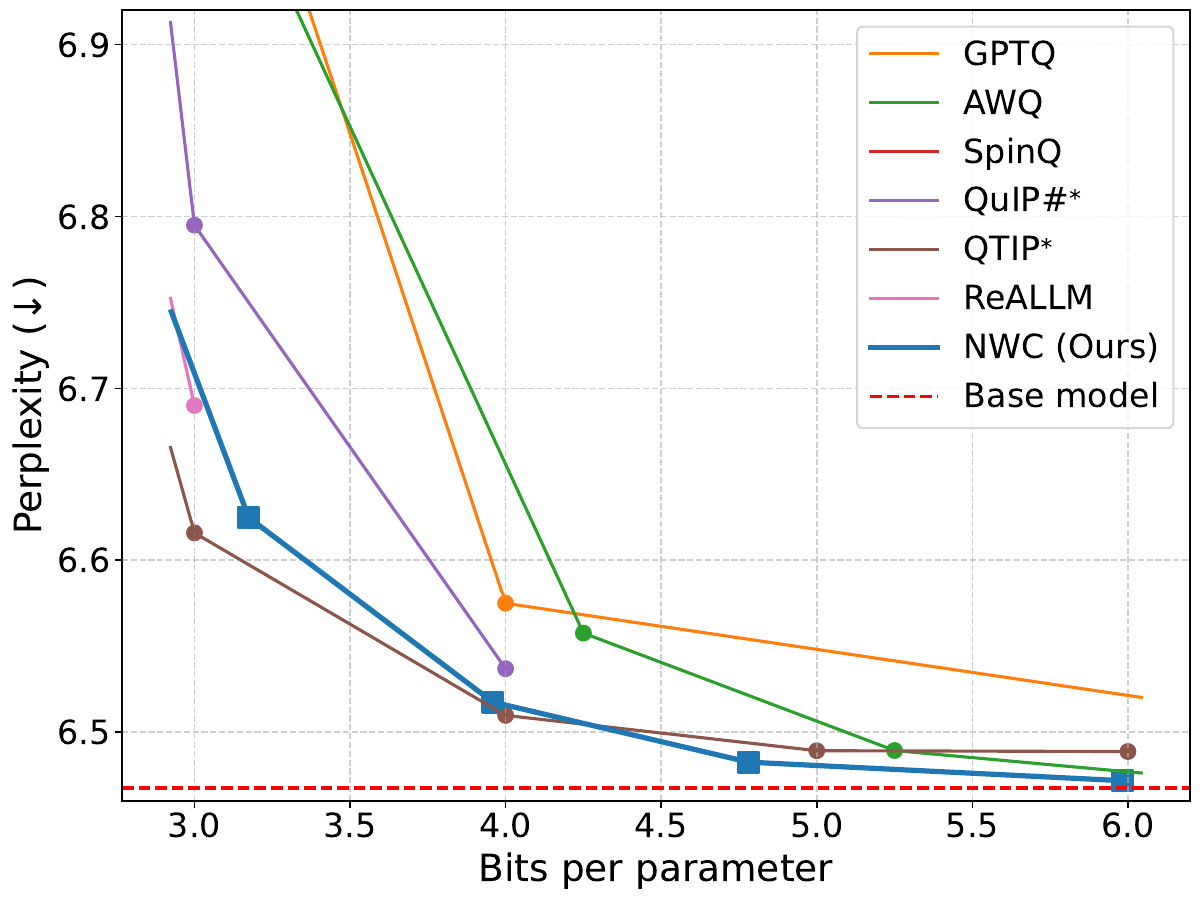}
        \caption{Llama 2-13B}
    \end{subfigure}
    \caption{Rate-accuracy performance on Llama models. We evaluate C4 perplexity (context length 2048) across various average bit-widths.}
    \label{fig:llama_c4}
\end{figure*}

%% file: figure_tex/result_other_models_common.tex
\begin{figure*}[t]
    \centering
    \hfill
    \begin{subfigure}[b]{0.32\textwidth}
        \includegraphics[width=\textwidth]{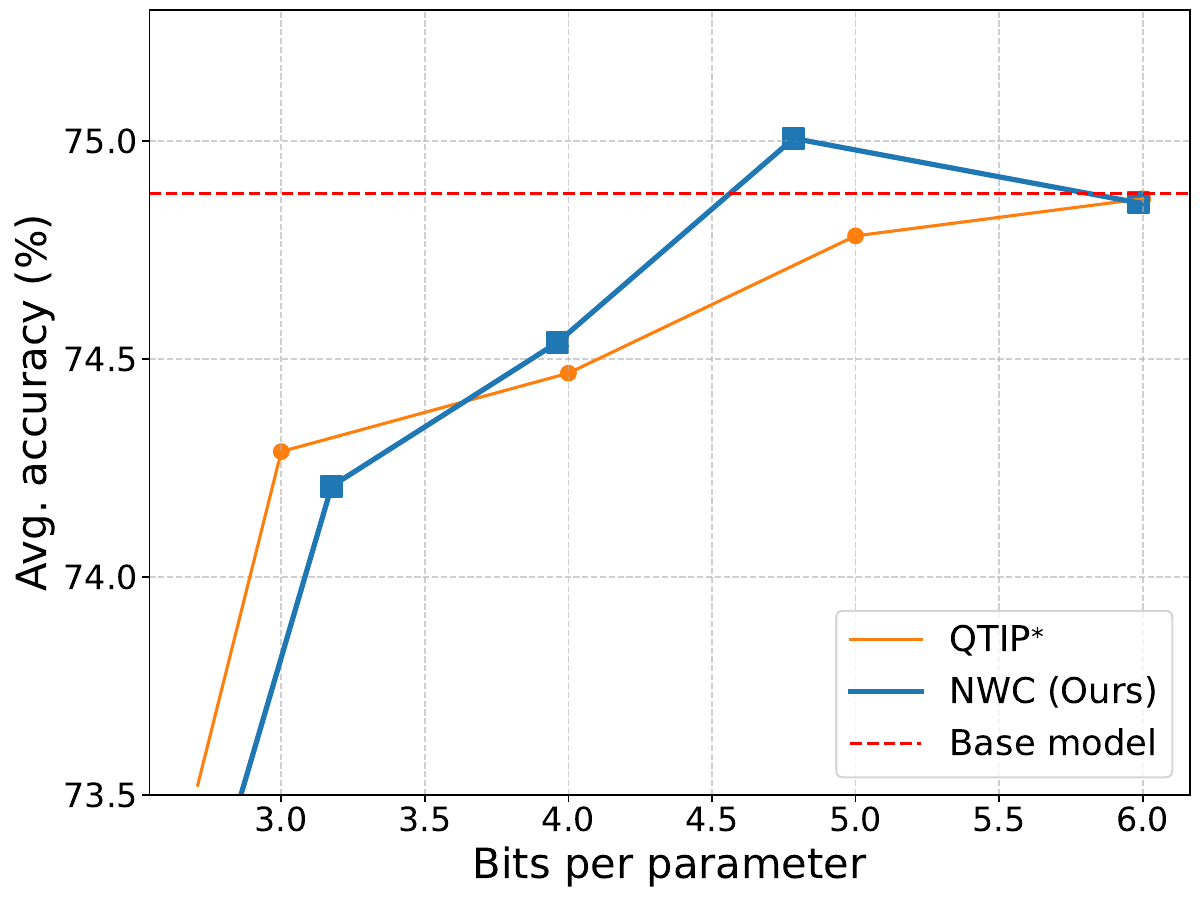}
        \caption{Mixtral-8x7B}
    \end{subfigure}
    \hfill
    \begin{subfigure}[b]{0.32\textwidth}
        \includegraphics[width=\textwidth]{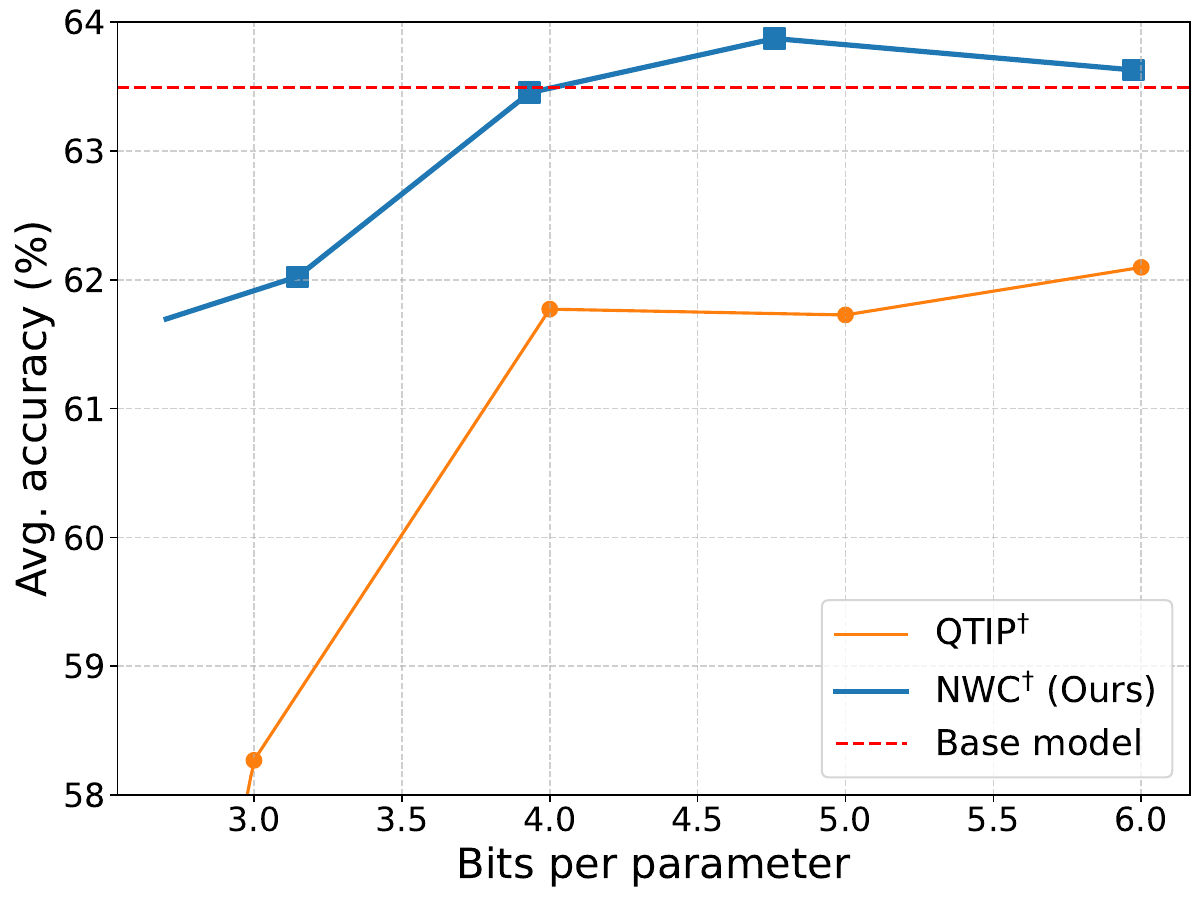}
        \caption{GPT-OSS-20B}
    \end{subfigure}
    \hfill
    \begin{subfigure}[b]{0.32\textwidth}
        \includegraphics[width=\textwidth]{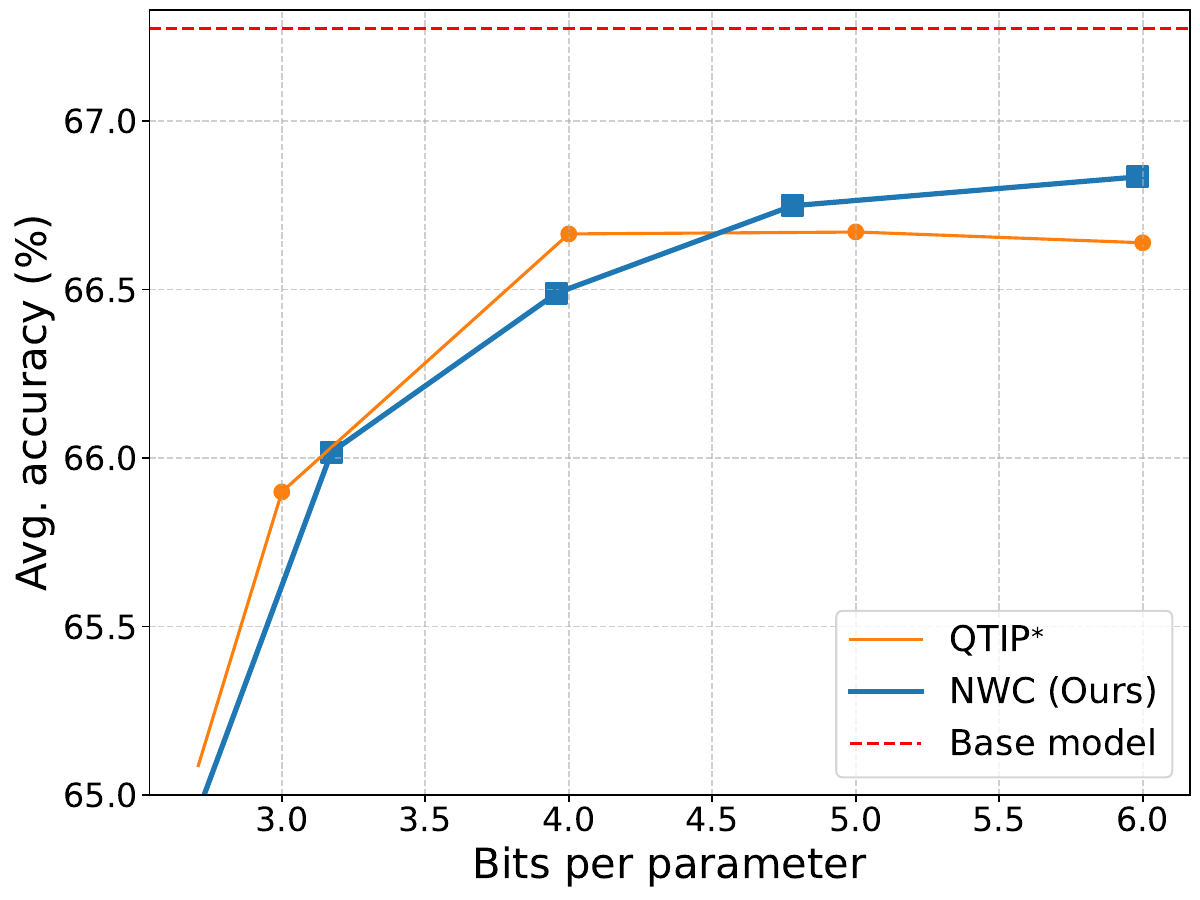}
        \caption{Llama 2-7B}
    \end{subfigure}
    \caption{Rate-accuracy tradeoffs on six common-sense tasks. We evaluate average zero-shot accuracies ARC-Easy, ARC-Challenge, WinoGrande, PiQA, HellaSwag, BoolQ across varying average bit-widths.}
    \label{fig:other_models_common}
\end{figure*}

%% file: figures/Llama2-7B_QAT.tex
\begin{figure*}[t]
    \centering
    \begin{subfigure}[b]{0.325\textwidth}
        \includegraphics[width=\textwidth]{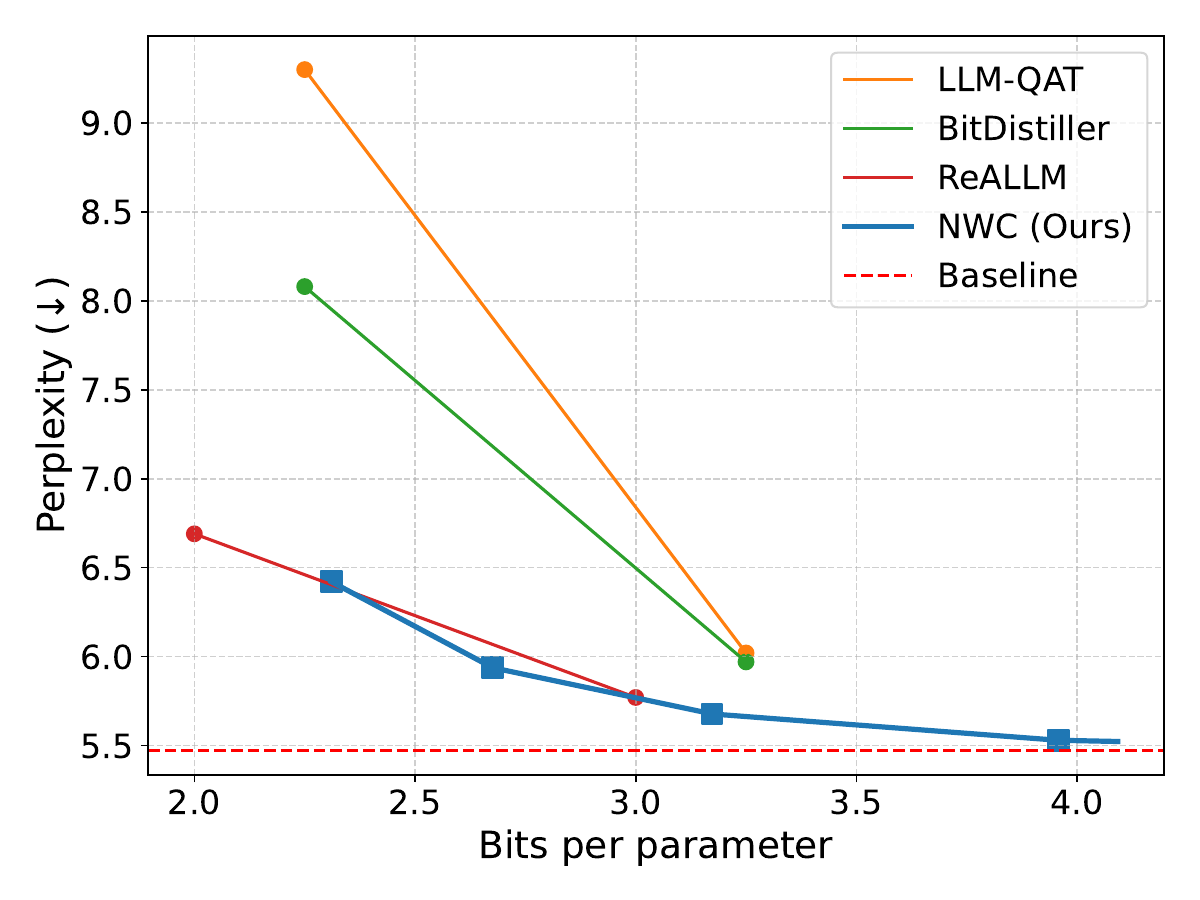}
        \caption{WikiText-2}
    \end{subfigure}
    \begin{subfigure}[b]{0.325\textwidth}
        \includegraphics[width=\textwidth]{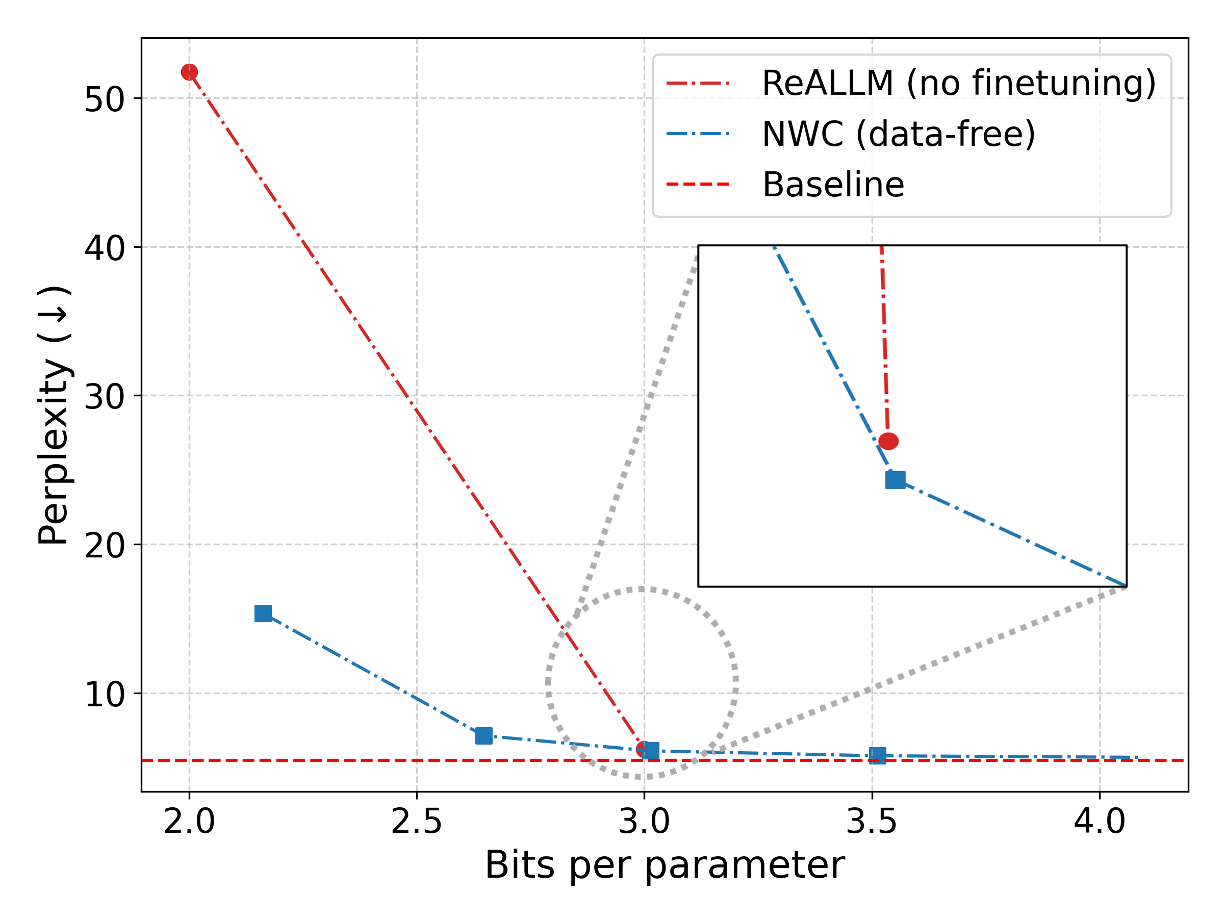}
        \caption{WikiText-2 (data-free)}
    \end{subfigure}
    \begin{subfigure}[b]{0.325\textwidth}
        \includegraphics[width=\textwidth]{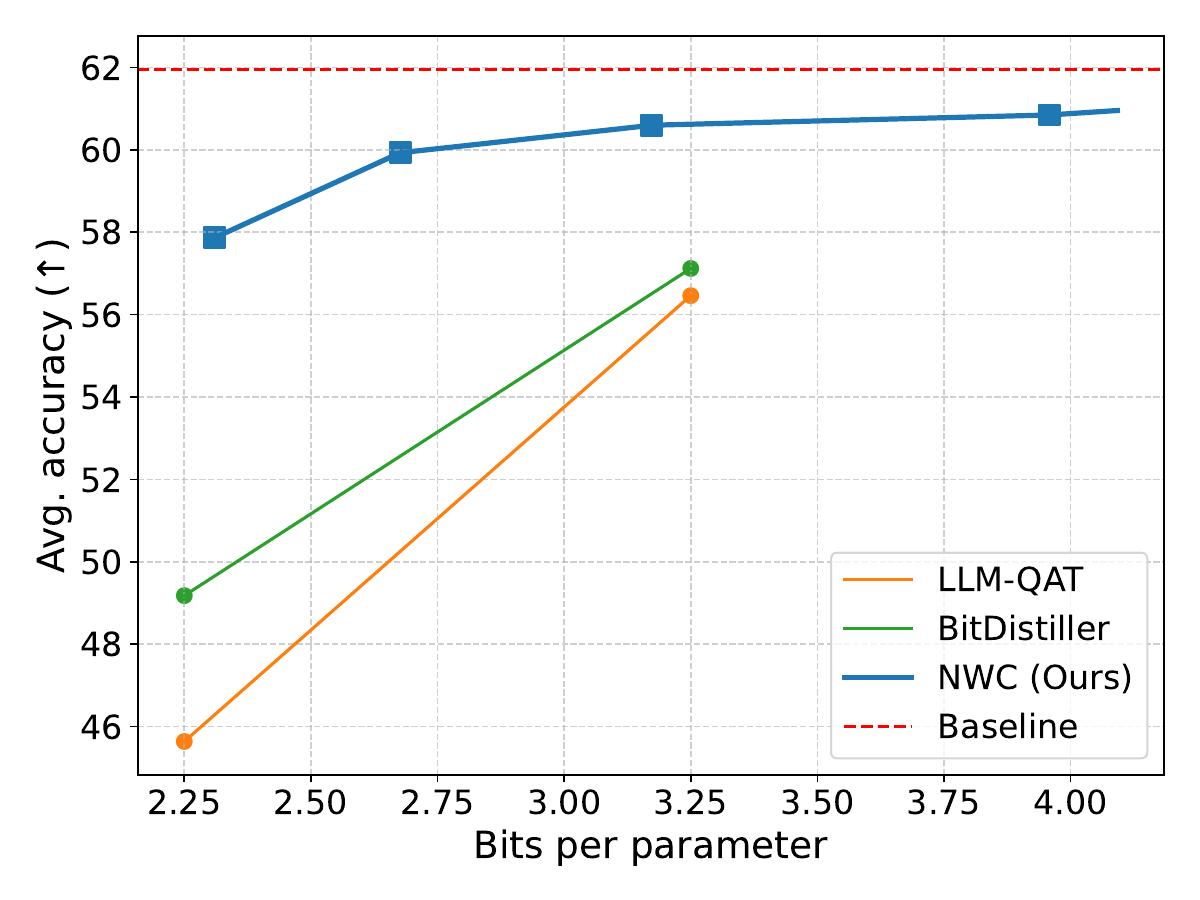}
        \caption{Zero-shot tasks}
    \end{subfigure}

    \caption{Compression results of Llama 2-7B.}
    \label{fig:comparision_qat}
\end{figure*}

%% file: tables/comparison_svd.tex
\begin{table*}[t]
\centering
\resizebox{0.7\textwidth}{!}{
\begin{tabular}{l c c c c c c c c c}
\toprule
\textbf{METHOD} & \textbf{Ratio}$\downarrow$ & \textbf{Wiki-2}$\downarrow$ & \textbf{OpenQ} & \textbf{ARCe} & \textbf{WG} & \textbf{HS} & \textbf{PQ} & \textbf{MathQ} & \textbf{Avg}$\uparrow$\\
\midrule
Original & 100\% & 6.14 & 0.35 & 0.80 & 0.73 & 0.60 & 0.80 & 0.40 & 0.61 \\
\midrule
SVD-LLM & 80\% & 11.82 & 0.29 & 0.77 & 0.64 & 0.51 & 0.72 & 0.30 & 0.54 \\
SVD-LLM V2 & 80\% & 8.01 & 0.33 & 0.79 & 0.70 & 0.58 & 0.77 & 0.36 & 0.59 \\
NWC        & 25\% & \textbf{6.32} & \textbf{0.34} & \textbf{0.80}  & \textbf{0.75} & \textbf{0.60} & \textbf{0.79} & \textbf{0.39} & \textbf{0.61} \\
\bottomrule
\end{tabular}
}
\caption{Performance of Llama 3-8B compressed by SVD-based methods and NWC.}\label{tab:comprasison_svd}
\end{table*}

%% file: tables/pocktllm.tex
\begin{table}[h]
    \centering
    \resizebox{0.4\textwidth}{!}{
    \begin{tabular}{lccc}
    \toprule
    \textbf{Method} & \textbf{Bits} & \textbf{WikiText-2} & \textbf{C4} \\
    \midrule
    FP16 & 16 & 5.12 & 6.63 \\
    \midrule
    PocketLLM & 3.98 & 5.27 & 6.86 \\
    NWC (Ours) & 3.96 & \textbf{5.17} & \textbf{6.70} \\
    \bottomrule
    \end{tabular}
    }
    \caption{Perplexity comparison of Llama 2-7B on WikiText-2 and C4 benchmarks (context length 4096).}\label{tab:pocketllm}
\end{table}

%% file: figures/ablation_main.tex
\begin{figure*}[t]
    \centering
    \begin{subfigure}[b]{0.49\textwidth}
        \includegraphics[width=\textwidth]{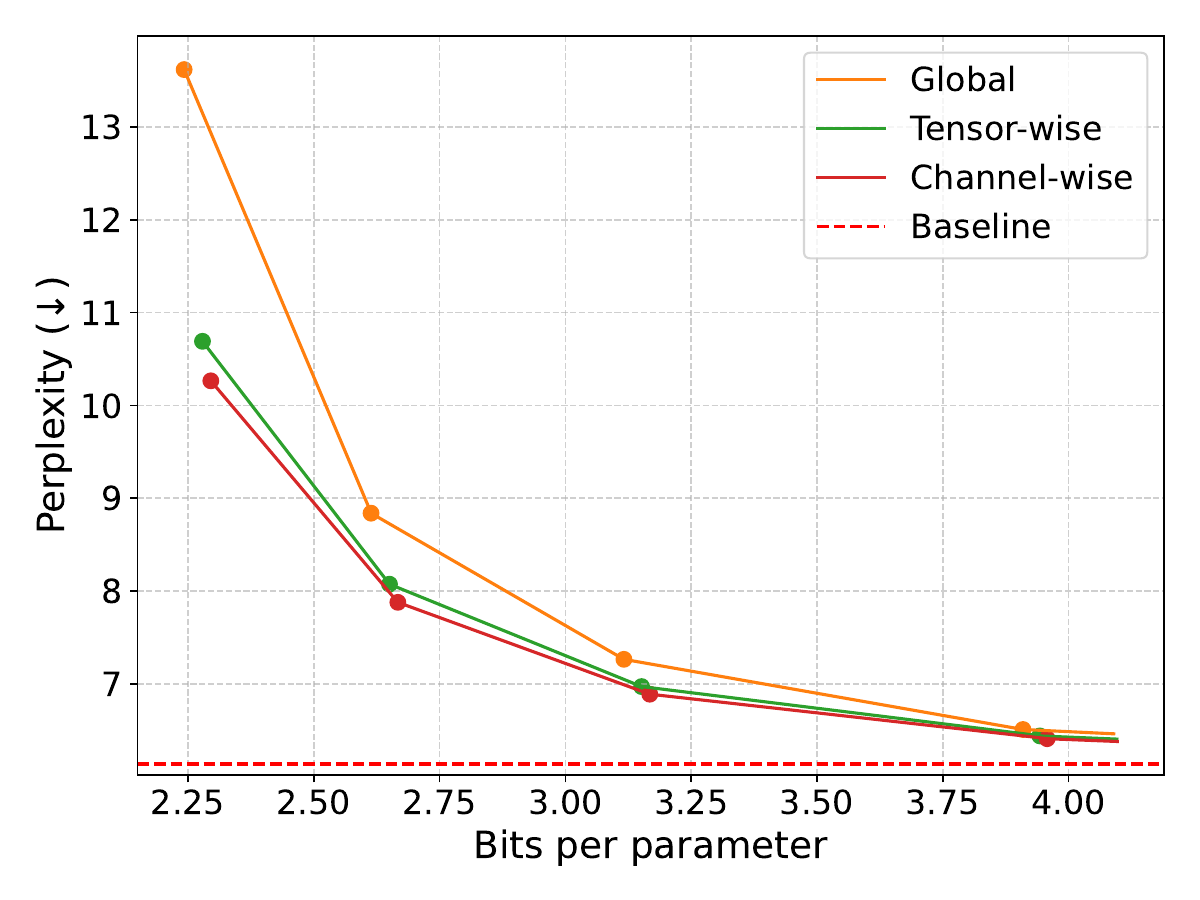}
        \caption{Normalization}
        \label{fig:abl_normalization}
    \end{subfigure}
    \begin{subfigure}[b]{0.49\textwidth}
        \includegraphics[width=\textwidth]{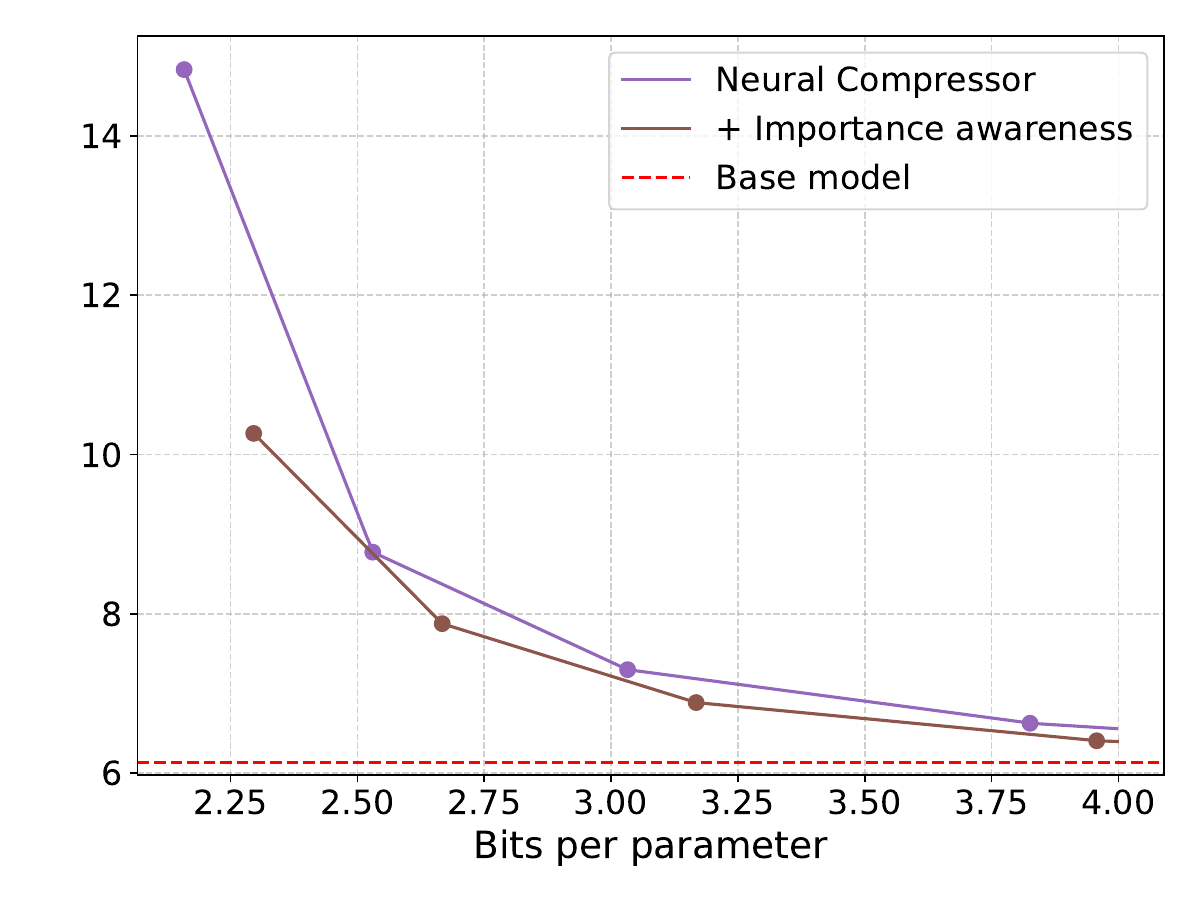}
        \caption{Importance-awareness}
        \label{fig:abl_importance}
    \end{subfigure}
    \begin{subfigure}[b]{0.49\textwidth}
        \includegraphics[width=\textwidth]{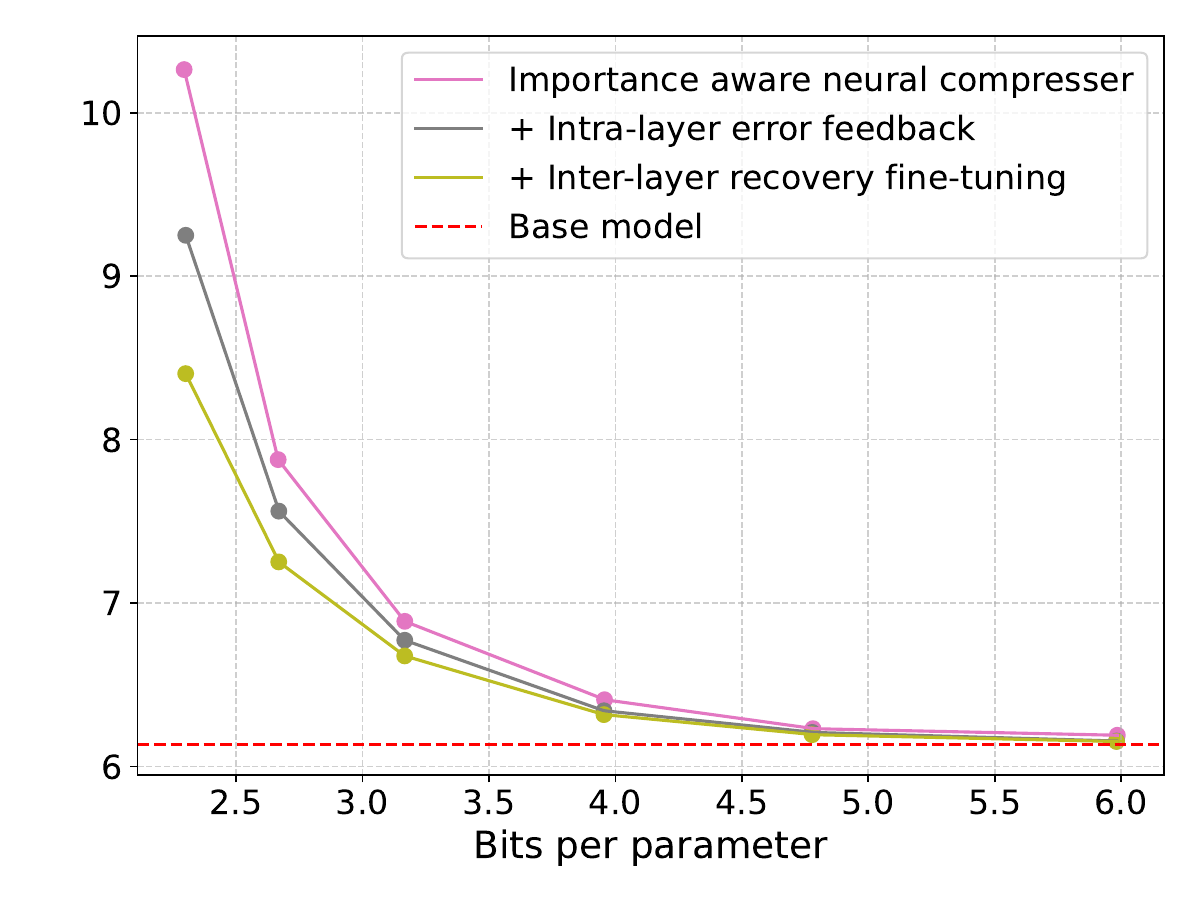}
        \caption{Error compensation}
        \label{fig:abl_eror}
    \end{subfigure}
    \begin{subfigure}[b]{0.49\textwidth}
        \includegraphics[width=\textwidth]{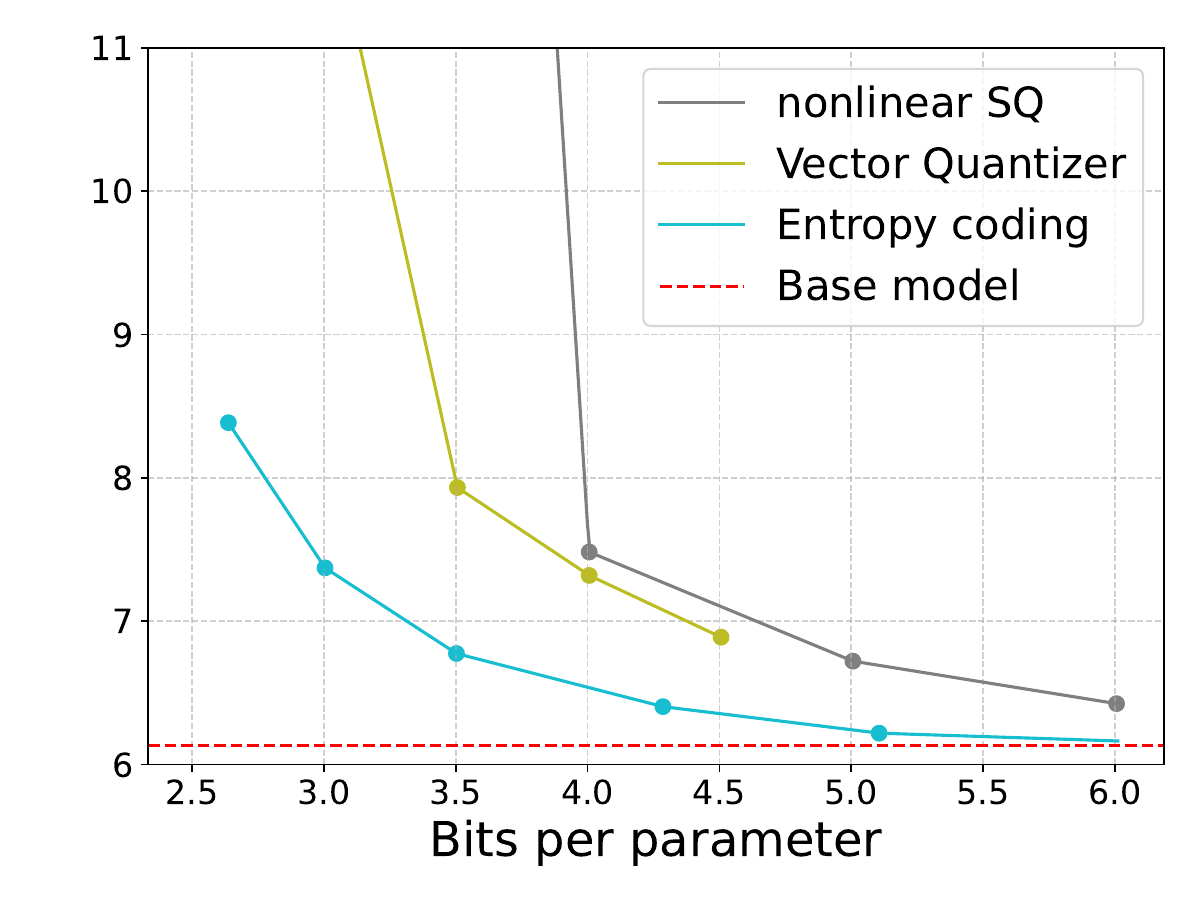}
        \caption{Entropy model}
        \label{fig:abl_entropy}
    \end{subfigure}
    \caption{Ablation studies on NWC components. We report WikiText-2 perplexity (context length 2048) on Llama 3-8B.}
    \label{fig:ablation_main}
\end{figure*}

%% file: tables/ablation_encoder.tex
\begin{table}[t]
\centering
\begin{tabular}{lccc}
\toprule
\textbf{En/Decoder} & \textbf{Bits} & \textbf{Wikitext-2$\downarrow$} & \textbf{C4$\downarrow$} \\
\midrule
$\mathrm{Identity}(\cdot)$ & 2.12 & 27.8 & 33.2 \\
$g_a / g_s$ & 2.13 & 14.5 & 20.7 \\
\bottomrule
\end{tabular}
\caption{Wikitext-2 and C4 perplexity for Llama-3-8B.}\label{tab:ablation_encoder}
\end{table}

%% file: tables/ablation_chunksize.tex
\begin{table}[tbp]
\centering
\resizebox{\columnwidth}{!}{%
\begin{tabular}{llccccc}
\toprule
\textbf{Dataset} & \textbf{CS} & \textbf{2.3bit} & \textbf{2.7bit} & \textbf{3.2bit} & \textbf{4.0bit} & \textbf{4.8bit} \\
\midrule
\textbf{WikiText-2} & \textbf{4}  & 9.06  & 7.52  & 6.77  & 6.34  & 6.20 \\
\textit{(Base: 6.14)} & \textbf{16} & 8.92  & 7.50  & 6.75  & 6.35  & 6.21 \\
                    & \textbf{64} & 9.15  & 7.54  & 6.79  & 6.37  & 6.21 \\
\midrule
\textbf{C4}         & \textbf{4}  & 12.54 & 10.76 & 9.72  & 9.12  & 8.95 \\
\textit{(Base: 8.88)} & \textbf{16} & 12.39 & 10.73 & 9.70  & 9.12  & 8.96 \\
                    & \textbf{64} & 12.56 & 10.81 & 10.07 & 9.15  & 8.96 \\
\bottomrule
\end{tabular}
}
\caption{Ablation on chunk size. We measure WikiText-2 perplexity (context length 2048) on Llama 3-8B.}\label{tab:ablation_chunksize}
\end{table}

%% file: figure_tex/ablation_K.tex
\begin{figure}[t]
    \centering
        \includegraphics[width=\columnwidth]{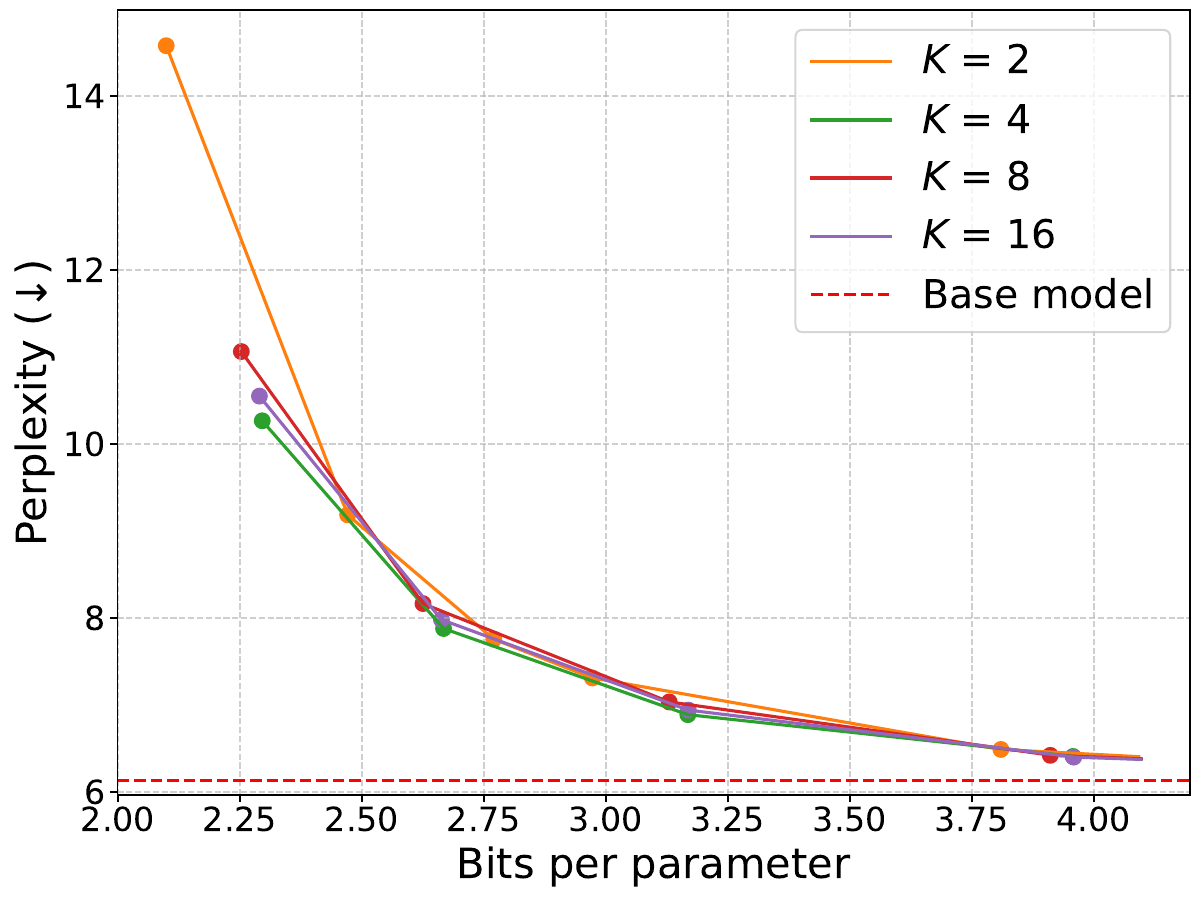}
    \caption{Ablation on the number of importance levels, $K$. We evaluate WikiText-2 perplexity with context length 2048}
    \label{fig:ablation_K}
\end{figure}

%% file: tables/ablation_hessian_assign.tex
\begin{table}[tbp]
\centering
\resizebox{\columnwidth}{!}{%
\begin{tabular}{llcccccc}
\toprule
\textbf{Dataset} & \textbf{Method} & \textbf{2.3b} & \textbf{2.7b} & \textbf{3.2b} & \textbf{4.0b} & \textbf{4.8b} & \textbf{6.0b} \\
\midrule
WikiText-2 & Random  & 19.26 & 9.68 & 7.82 & 6.56 & 6.29 & 6.19 \\
           & Hessian & \textbf{10.27} & \textbf{7.88} & \textbf{6.89} & \textbf{6.40} & \textbf{6.22} & \textbf{6.17} \\
\midrule
C4         & Random  & 26.06 & 13.52 & 10.36 & 9.39 & 9.07 & \textbf{8.93} \\
           & Hessian & \textbf{14.07} & \textbf{11.28} & \textbf{9.91} & \textbf{9.20} & \textbf{8.99} & \textbf{8.93} \\
\bottomrule
\end{tabular}
}
\caption{Ablation on Hessian-based importance assignment}\label{tab:ablation_hess}
\end{table}

%% file: figures/layer_stats.tex
\begin{figure}[t]
    \centering
    \begin{subfigure}[b]{\columnwidth}
        \includegraphics[width=\columnwidth]{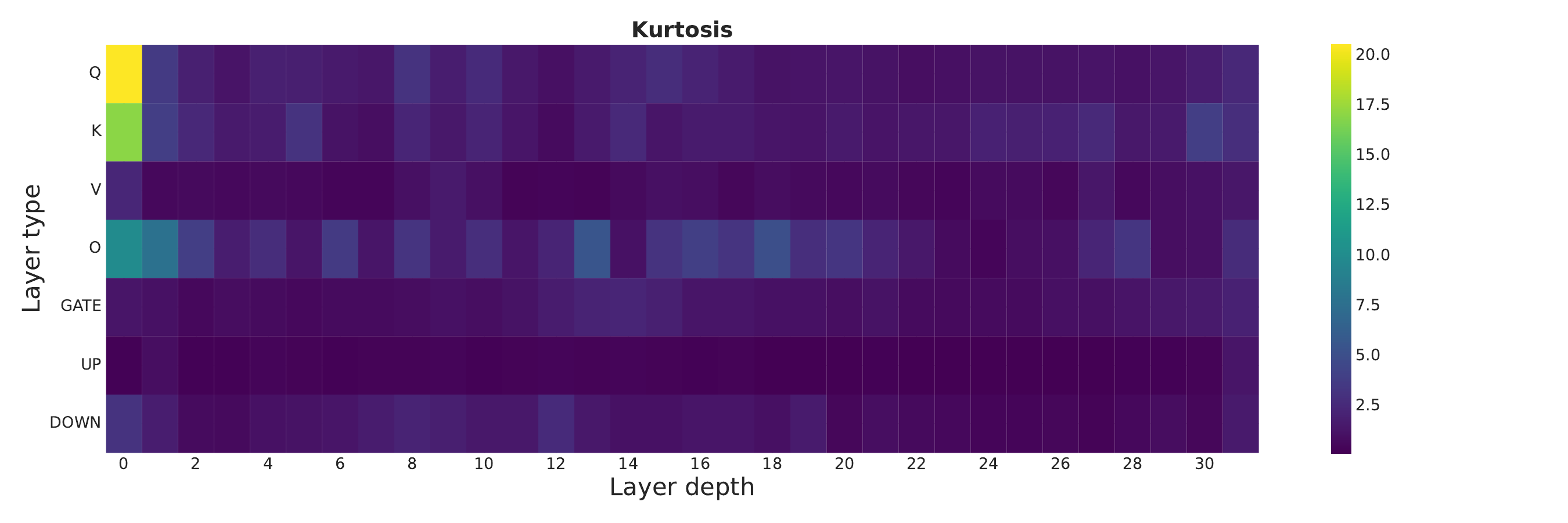}
    \end{subfigure}
    \begin{subfigure}[b]{\columnwidth}
        \includegraphics[width=\columnwidth]{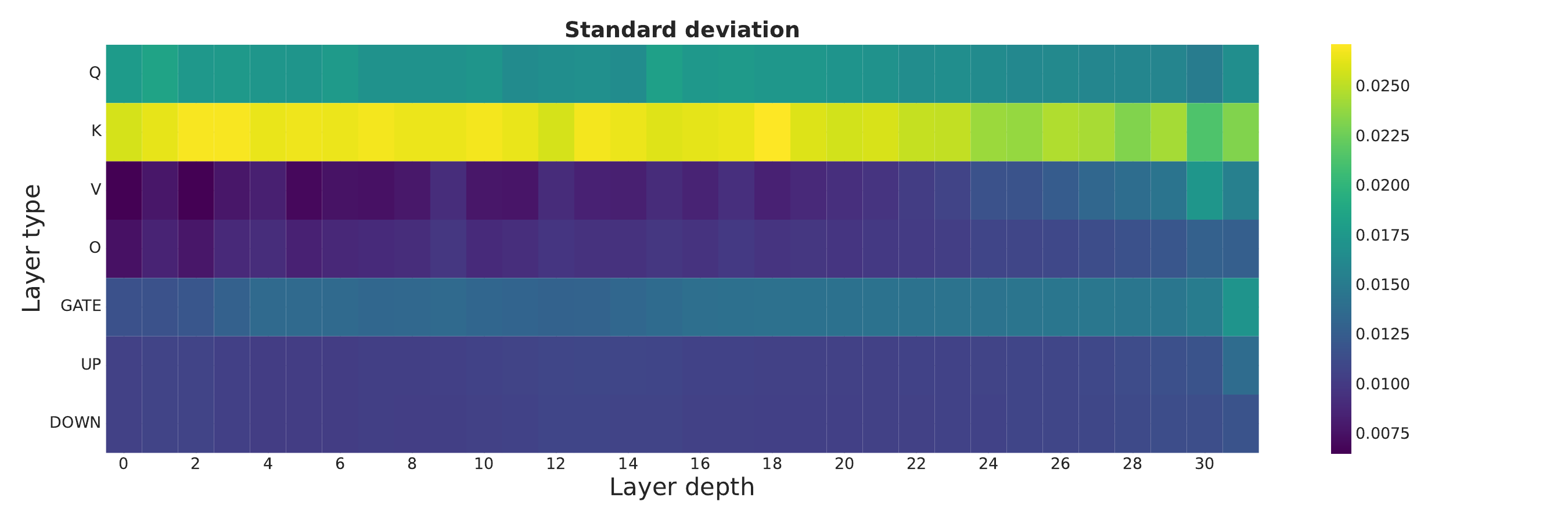}
    \end{subfigure}
    \caption{Layer-wise statistic of Llama 3-8B. (Top) Kurtosis. (Bottom) Standard deviation. }
    \label{fig:layer_stats}
\end{figure}

%% file: figure_tex/weight_channel_box.tex
\begin{figure}[t]
    \centering
    \includegraphics[width=\columnwidth]{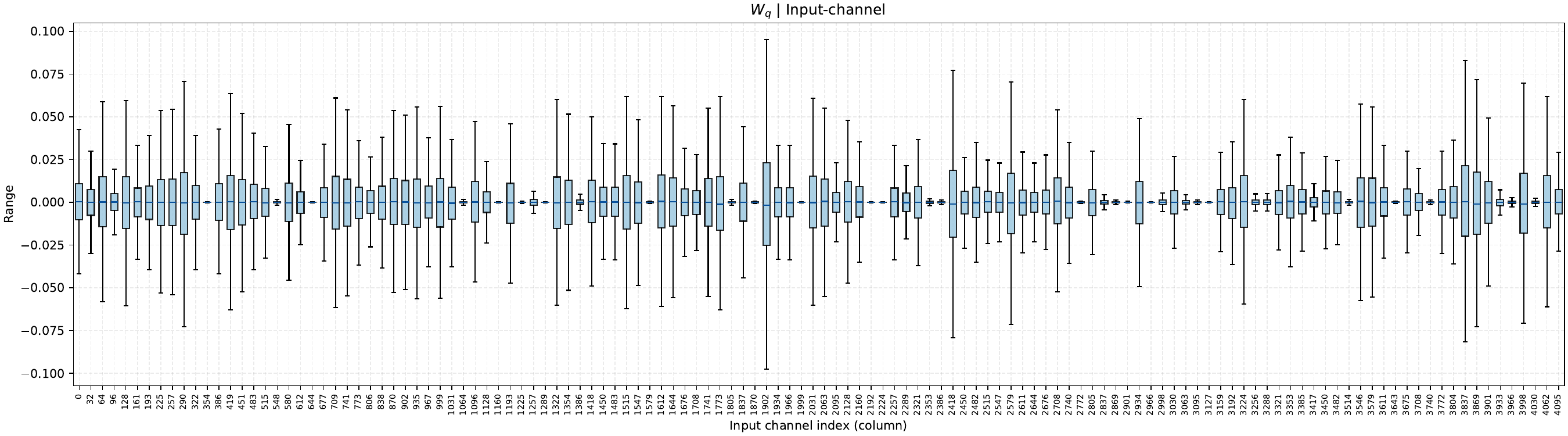}
    \caption{Channel-wise range of Q projection of Llama 3-8B.}
    \label{fig:channel_box}
    \vspace{-1em}
\end{figure}

%% file: figures/layerwise_RD.tex
\begin{figure}[t]
    \centering
    \begin{subfigure}[b]{0.49\columnwidth}
        \includegraphics[width=\columnwidth]{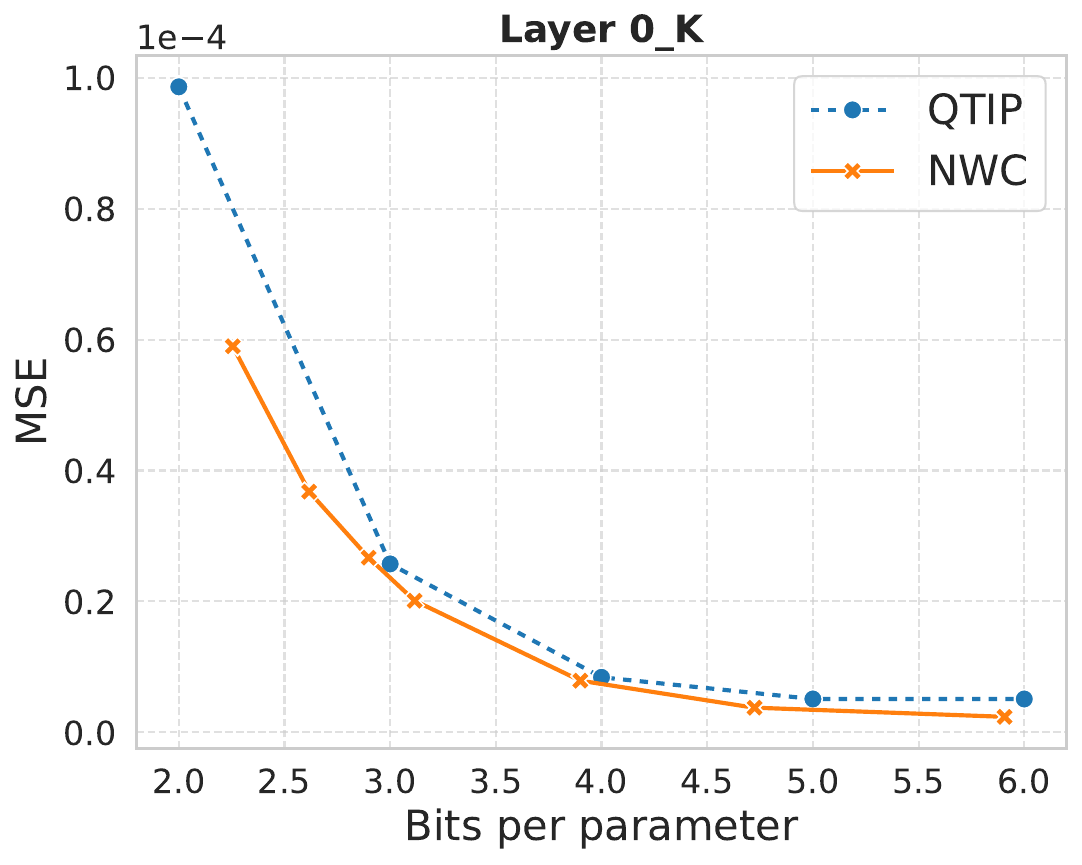}
    \end{subfigure}
    \begin{subfigure}[b]{0.49\columnwidth}
        \includegraphics[width=\columnwidth]{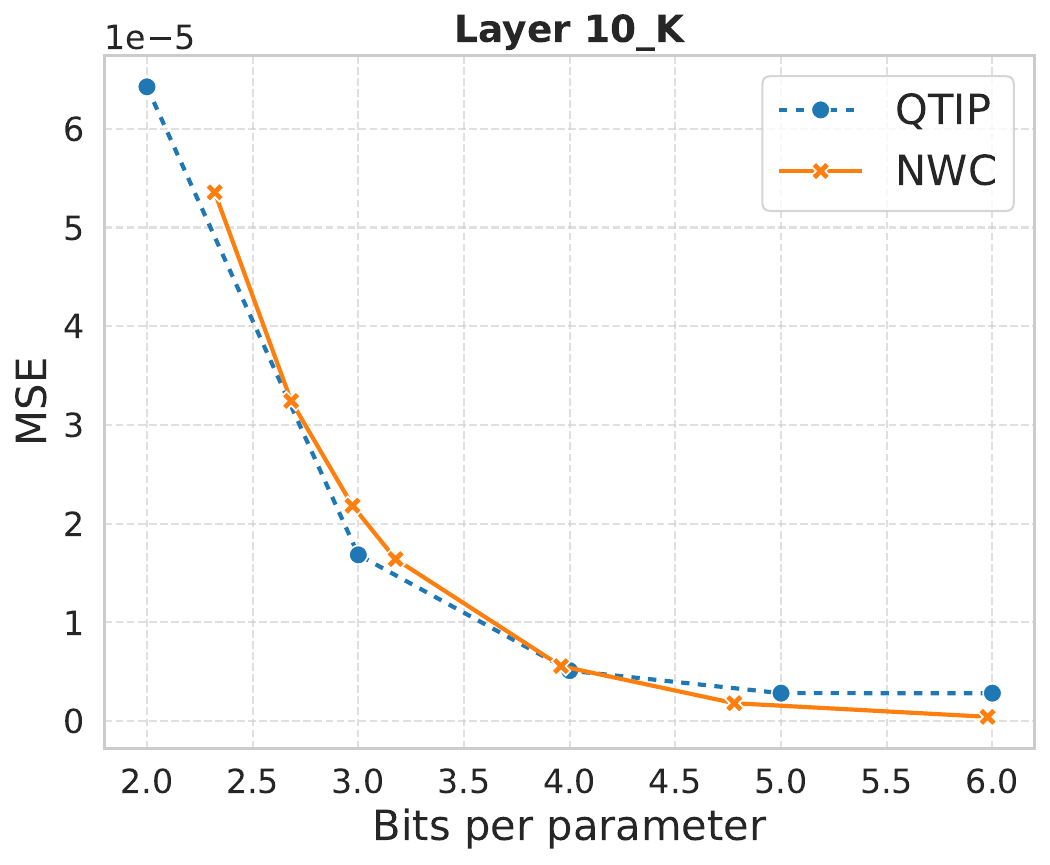}
    \end{subfigure}
    \caption{Per-layer rate-distortion curve of K projection layer in different blocks.}
    \label{fig:layerwise_rd}
\end{figure}

%% file: tables/8b_all_results.tex
\begin{table*}[h]
\centering

\begin{tabular}{c cc ccccccc}
\toprule
\multirow{2}{*}{Bit} & \multicolumn{2}{c}{Perplexity $\downarrow$} & \multicolumn{7}{c}{Zero-shot Accuracy (\%) $\uparrow$} \\
\cmidrule(lr){2-3} \cmidrule(lr){4-10}
 & Wiki & C4 & MMLU & ARC-C & ARC-E & BoolQ & PIQA & WG & HS \\
\midrule
2.30 & 8.39 & 11.72 & 51.07 & 41.15 & 74.27 & 80.42 & 76.66 & 70.72 & 53.01 \\
2.67 & 7.27 & 10.43 & 55.99 & 44.97 & 78.30 & 79.75 & 77.51 & 72.30 & 56.26 \\
3.17 & 6.68 &  9.59 & 59.97 & 48.35 & 79.91 & 81.42 & 79.16 & 74.09 & 58.46 \\
3.96 & 6.32 &  9.09 & 61.65 & 50.46 & 80.09 & 81.06 & 79.14 & 73.98 & 59.67 \\
4.80 & 6.19 &  8.94 & 61.95 & 50.77 & 80.40 & 80.97 & 79.67 & 73.90 & 59.96 \\
5.98 & 6.15 &  8.89 & 62.13 & 50.34 & 80.11 & 81.65 & 79.49 & 73.13 & 60.22 \\
\bottomrule
\end{tabular}
\caption{Llama3-8B results corresponding to \Cref{fig:8b_wiki_mmlu_common}, including perplexity on WikiText-2 and C4 as well as zero-shot accuracies on MMLU and common-sense benchmarks.}
\label{tab:nwc_results}
\end{table*}

\begin{table*}[h]
\centering
\begin{tabular}{cc cc cc}
\toprule
\multicolumn{2}{c}{Mixtral} & \multicolumn{2}{c}{Qwen3-30B-A3B} & \multicolumn{2}{c}{GPT-OSS 20B} \\
\cmidrule(lr){1-2} \cmidrule(lr){3-4} \cmidrule(lr){5-6}
Bit & MMLU (\%) & Bit & MMLU (\%) & Bit & MMLU (\%) \\
\midrule
2.31 & 61.51 & 2.31 & 72.38 & 2.24 & 30.51 \\
2.67 & 65.13 & 2.67 & 75.06 & 2.62 & 43.92 \\
3.17 & 66.55 & 3.17 & 76.10 & 3.11 & 50.39 \\
3.95 & 67.69 & 3.95 & 77.17 & 3.90 & 51.94 \\
4.78 & 67.96 & 4.78 & 77.38 & 4.75 & 53.76 \\
5.95 & 68.15 & 5.95 & 77.65 & 5.94 & 52.05 \\
\bottomrule
\end{tabular}
\caption{MMLU results for other LLM architectures shown in \Cref{fig:other_models}}
\label{tab:nwc_mmlu}
\end{table*}

\begin{table*}[h]
\centering
\begin{tabular}{cc cc cc}
\toprule
\multicolumn{2}{c}{SigLIP} & \multicolumn{2}{c}{CLIP-L/14} & \multicolumn{2}{c}{DINOv2-L} \\
\cmidrule(lr){1-2} \cmidrule(lr){3-4} \cmidrule(lr){5-6}
Bit & MMLU (\%) & Bit & MMLU (\%) & Bit & MMLU (\%) \\
\midrule
2.32 & 50.27 & 2.30 & 50.79 & 2.31 & 84.38 \\
2.69 & 60.56 & 2.67 & 61.99 & 2.67 & 85.35 \\
3.18 & 66.20 & 3.16 & 67.69 & 3.16 & 85.81 \\
3.97 & 68.85 & 3.95 & 69.94 & 3.95 & 86.02 \\
4.79 & 69.48 & 4.77 & 70.83 & 4.77 & 86.07 \\
5.98 & 69.64 & 5.94 & 71.10 & 5.96 & 86.09 \\
\bottomrule
\end{tabular}
\caption{ImageNet Top-1 accuracy results for the vision encoders shown in \Cref{fig:vision_models}}
\label{tab:vision_results}
\end{table*}